%% file: main.tex
\documentclass[11pt]{article}

\usepackage[final]{acl}

\usepackage{times}
\usepackage{setspace}
\usepackage{latexsym}
\usepackage{booktabs}

\usepackage[T1]{fontenc}

\usepackage[utf8]{inputenc}

\usepackage{microtype}

\usepackage{inconsolata}

\usepackage{graphicx}
\usepackage{amsmath}
\usepackage{subcaption}
\usepackage{amssymb}
\usepackage{multirow}
\usepackage{colortbl}
\usepackage{enumitem}
\usepackage[table,xcdraw]{xcolor}
\usepackage[ruled,vlined,linesnumbered]{algorithm2e}

\usepackage[most]{tcolorbox}
\definecolor{agentcolor}{RGB}{255, 255, 255}
\newtcolorbox{dialoguebox}[3][]{
    breakable, 
    enhanced,
    before skip=3pt, 
    after skip=3pt,
    left=3pt, right=3pt, top=3pt, bottom=3pt,
    colframe=#2!60!black, 
    colback=#2,
    title={\textbf{#3}}, 
    fonttitle=\sffamily\tiny,
    fontupper=\footnotesize, 
    before upper={\setstretch{1.05}}, 
    arc=2pt,
    boxrule=1.5pt,
    parbox=false, 
    #1
}

\title{GUITester: Enabling GUI Agents for Exploratory Defect Discovery}

\author{
  \textbf{Yifei Gao\textsuperscript{1}\thanks{Work done during internship at Hithink Research.}},
  \textbf{Jiang Wu\textsuperscript{2}},
  \textbf{Xiaoyi Chen\textsuperscript{1}},
  \textbf{Yifan Yang\textsuperscript{1}},
\\
  \textbf{Zhe Cui\textsuperscript{2}},
  \textbf{Tianyi Ma\textsuperscript{2}},
  \textbf{Jiaming Zhang\textsuperscript{3}},
  \textbf{Jitao Sang\textsuperscript{1}*}
\\
\\
  \textsuperscript{1}Beijing Jiaotong University,
  \textsuperscript{2}Hithink Research,
  \textsuperscript{3}Nanyang Technological University
\\
  \small{
    \textbf{Correspondence:} \href{mailto:jtsang@bjtu.edu.cn}{jtsang@bjtu.edu.cn}
  }
}

\begin{document}
\maketitle

\begin{figure*}[tp]
    \centering
    \includegraphics[width=1.0\linewidth]{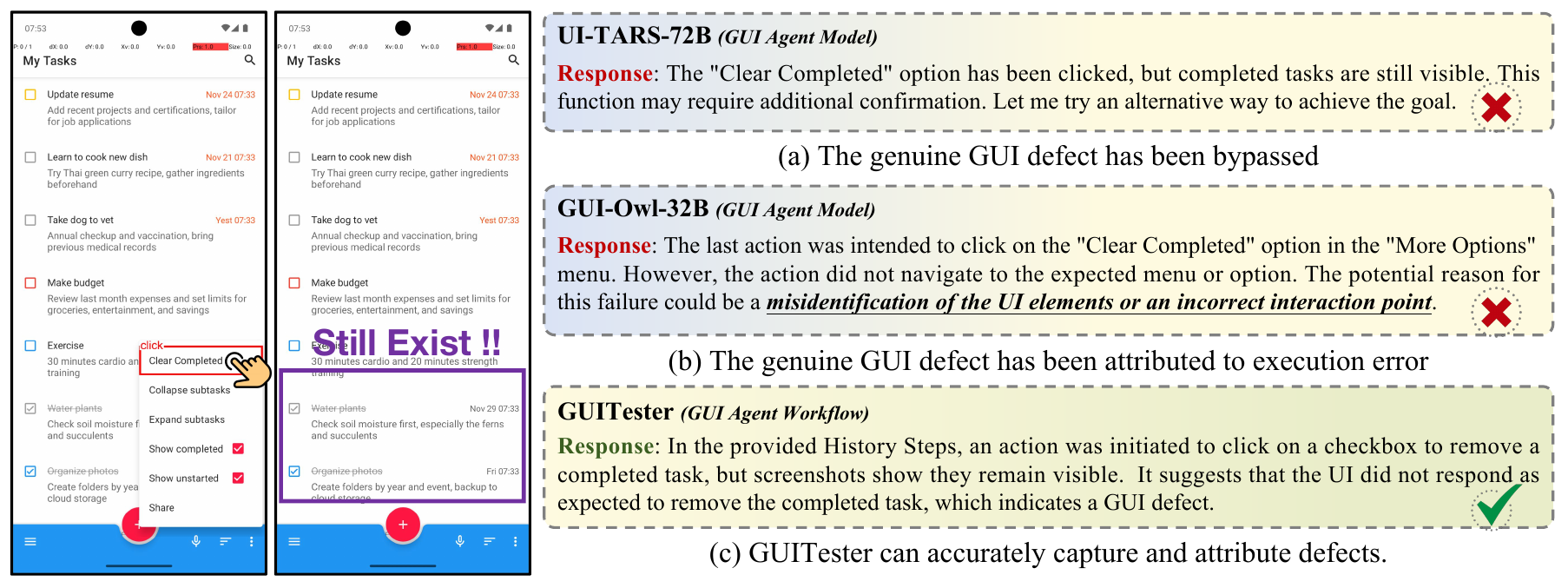}
    \caption{A GUI defect where "Clear Completed" fails to remove the completed tasks. Existing approaches either bypass the defect by attempting alternative paths, or misattribute it to agent execution errors, while GUITester accurately captures the anomaly and attributes it as a genuine GUI defect.}
    \label{fig:new_intro}
\end{figure*}

\begin{abstract}
Exploratory GUI testing is essential for software quality but suffers from high manual costs. While Multi-modal Large Language Model (MLLM) agents excel in navigation, they fail to autonomously discover defects due to two core challenges: \textit{Goal-Oriented Masking}, where agents prioritize task completion over reporting anomalies, and \textit{Execution-Bias Attribution}, where system defects are misidentified as agent errors. To address these, we first introduce \textbf{GUITestBench}, the first interactive benchmark for this task, featuring 143 tasks across 26 defects. We then propose \textbf{GUITester}, a multi-agent framework that decouples navigation from verification via two modules: (i) a \textit{Planning-Execution Module (PEM)} that proactively probes for defects via embedded testing intents, and (ii) a \textit{Hierarchical Reflection Module (HRM)} that resolves attribution ambiguity through interaction history analysis. GUITester achieves an F1-score of 48.90\% (Pass@3) on GUITestBench, outperforming state-of-the-art baselines (33.35\%). Our work demonstrates the feasibility of autonomous exploratory testing and provides a robust foundation for future GUI quality assurance~\footnote{Our code is now available in~\href{https://github.com/ADaM-BJTU/GUITestBench}{https://github.com/ADaM-BJTU/GUITestBench}}.

\end{abstract}

\section{Introduction}

Exploratory GUI testing is a critical paradigm for ensuring software reliability by uncovering defects within unscripted, complex interaction contexts \citep{Survey1, Survey2}. Unlike script-based testing \citep{TestScript}, it inherently requires an autonomous navigation of the interface and detection of defects without predefined test oracles \citep{intro-1}. However, the efficacy of this methodology is traditionally bottlenecked by its heavy reliance on human expertise and subjective judgment, which precludes large-scale, continuous execution in modern rapid-development cycles \citep{intro-2}. While Multimodal Large Language Model (MLLM)-powered GUI agents have demonstrated remarkable proficiency in GUI navigation \citep{UI-TARS, mobile-agent-v3,MAIUI}, their potential for autonomous defect discovery remains largely unfulfilled.

We identify two fundamental challenges that prevent existing GUI agents from effective exploratory testing: \textbf{(i) Goal-Oriented Masking.} Most GUI agents are optimized to maximize task success rates, which inherently encourages robustness against environmental obstacles. In a testing context, this goal-oriented nature leads the agent to perceive functional anomalies as traversable hurdles rather than reportable defects. As shown in Figure \ref{fig:new_intro}(a), when encountering a non-responsive button, the agent’s policy autonomously seeks alternative navigation paths to reach the goal. This ``success-at-all-costs'' behavior effectively masks the defect, rendering it invisible to the quality assurance pipeline.
\textbf{(ii) Execution-Bias Attribution.} Exploratory testing lacks explicit oracles, requiring agents to distinguish between their own operational failures (e.g., coordinate miscalculations) and genuine software defects. Due to the stochastic nature of MLLM interactions, current agents exhibit a systematic bias toward self-attribution: erroneously assuming that any failure to trigger a state change stems from their own execution imprecision. As illustrated in Figure \ref{fig:new_intro}(b), GUI-Owl misinterprets a system-level rendering failure as a misaligned click, causing the genuine defect to be misclassified as a transient execution error in the logs.

To investigate the capability of existing GUI agents in exploratory testing scenarios, we introduce \textbf{GUITestBench}, the first interactive benchmark for exploratory GUI defect detection. We collect 26 real-world defects across 12 Android applications and construct 143 navigation tasks that encounter these defects during execution. Given defect-agnostic task descriptions, agents must autonomously discover and report defects. To address the aforementioned challenges, we propose \textbf{GUITester}, a multi-agent framework that decouples navigation from defect verification. GUITester employs a \textbf{Planning Execution Module (PEM)} that intentionally probes for potential failures, preventing defects from being overlooked by goal-oriented navigation. It further introduces a \textbf{Hierarchical Reflection Module (HRM)} that utilizes interaction history to resolve the attribution dilemma, ensuring that software-side defects are not misattributed to agent-side execution slips.

We evaluate GUITester against state-of-the-art GUI automation agents on GUITestBench, including UI-TARS~\cite{UI-TARS}, GUI-Owl~\cite{mobile-agent-v3}, and MAI-UI~\cite{MAIUI}. Our results show that existing agents struggle with defect discovery, with the strongest baseline achieving only 33.35\% F1-score (Pass@3). In contrast, GUITester significantly improves the F1-score to 48.90\%, demonstrating the viability of autonomous exploratory testing and informing future agent design. Our contributions include: 
\begin{itemize}[leftmargin=1em, itemsep=-2pt]
\item We define the challenges of task-success bias and attribution ambiguity in MLLM-driven exploratory GUI testing. 
\item We introduce GUITestBench, an interactive evaluation framework featuring diverse, real-world GUI defects. 
\item We propose GUITester, a multi-agent framework featuring proactive defect probing and hierarchical reflection, which significantly improves defect discovery rates.
\end{itemize}

\begin{figure*}[tp]
    \centering
    \includegraphics[width=1.0\linewidth]{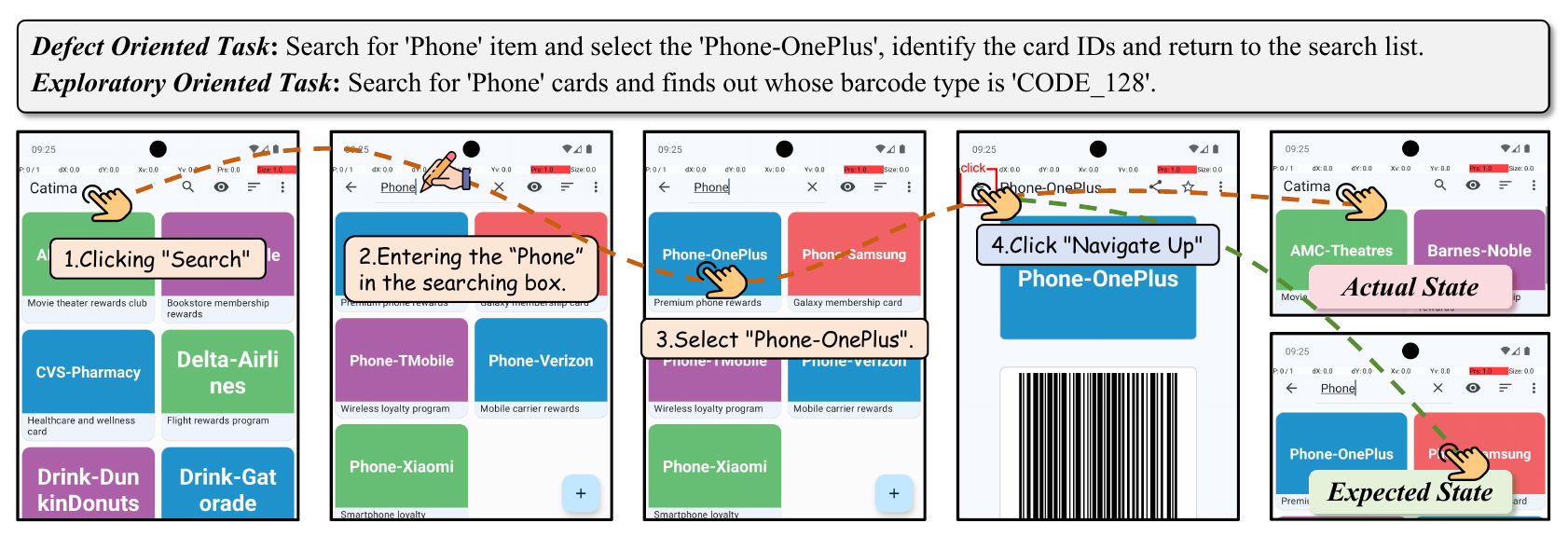}
    \caption{Example of a Navigation Logic Error defect with two task types. The defect-oriented task explicitly guides the agent to trigger the defect, while the exploratory-oriented task may encounter it during exploration. After clicking "Navigate Up", the app returns to the home page (actual) instead of the searching list (expected).}
    \label{fig:guitestbench_case}
\end{figure*}

\section{Related Works}

\noindent\textbf{GUI Agent.} Recent advances in multimodal large language models have enabled GUI agents to perform autonomous navigation. Agent workflows such as Mobile-Agent~\cite{mobile-agent-v1,mobile-agent-v2,mobile-agent-v3} leverage multi-agent collaboration and memory mechanisms for complex multi-step tasks, while AppAgent~\cite{AppAgent} incorporates document-augmented exploration learning. Meanwhile, Agent models like UI-TARS~\cite{UI-TARS,UI-TARS2} provide end-to-end GUI interaction capabilities, and GUI-Owl~\cite{mobile-agent-v3}, MAI-UI~\cite{MAIUI} enhances navigation through online reinforcement learning. These developments indicate that GUI agents have acquired fundamental capabilities for autonomous navigation, laying the groundwork for downstream applications such as test automation.

\noindent\textbf{Test Automation.} Several approaches have applied large language models to GUI testing. AUITestAgent~\cite{AUITestAgent}, GUIPilot~\cite{GUIPilot}, and ProphetAgent~\cite{ProphetAgent} automate specific testing workflows such as test case execution and consistency validation. Temac~\cite{Temac} further introduces multi-agent collaboration for testing tasks. While these approaches show promising results in component-centric testing scenarios, they still rely on predefined test cases. In contrast, exploratory testing, where human testers autonomously navigate applications to discover latent defects without predetermined intents, as in monkey testing, remains largely untouched by GUI agents, being one of the most time-consuming and difficult-to-scale aspects of manual testing.

\noindent\textbf{Benchmarks for GUI Testing.} Existing benchmarks, such as GTArena~\cite{GTArena}, focus on evaluating the capabilities of general models in static testing scenarios; however, fail to capture the dynamic nature of real-world applications. There remains a lack of benchmarks specifically designed to evaluate GUI agents' ability to autonomously discover defects through exploration. To fill this gap, we introduce GUITestBench, the first benchmark for evaluating exploratory GUI testing. 

\section{GUITestBench}
GUITestBench is a benchmark that enables agents to interact with mobile apps and discover defects through multi-step operations.  This benchmark evaluates three core capabilities for exploratory GUI testing: navigating to defect locations, recognizing anomalous behaviors, and reporting identified defects. We detail the construction process in §\ref{GTBench_construction} and the evaluation methodology in §\ref{GTBench_evaluation}.

\subsection{Benchmark Construction}\label{GTBench_construction}
We construct GUITestBench through a two-stage process: collecting and categorizing real-world defects (§\ref{GTBench_defect_cate}), and synthesizing exploratory tasks with controlled guidance levels (§\ref{GTBench_task_generation}). Dataset statistics are summarized in §\ref{GTBench_data_statistics}.

\subsubsection{Defect Collection and Categorization}\label{GTBench_defect_cate}
We collect GUI defects from public issues on GitHub, and these defects are categorized into UI functional defects and user experience (UX) defects. Specifically, UI defects stem from implementation errors in specific components, manifesting as failures in the expected functionality of elements (e.g., click failures or incorrect navigation), while UX defects~\cite{UXDebt} originate from design flaws in the interaction logic or task flow of multiple components, manifesting as anomalies in the interaction process, which usually cannot be attributed to a single faulty component.

We adopt three fault modes from the defect categories defined in GTArena~\cite{GTArena}: (1) \textbf{Operation No Response (ONR)}: an interaction yields no observable feedback, which applies only to UI defects, as unresponsiveness can be directly attributed to a specific element; (2) \textbf{Unexpected Task Result (UTR)}: the outcome deviates from expectations; (3) \textbf{Navigation Logic Error (NLE)}: flawed logic causes incorrect navigation flow. More examples are provided in Appendix~\ref{appsec:GUITestBench_cases}.

\subsubsection{Exploratory Task Synthesis}\label{GTBench_task_generation}
In practical exploratory GUI testing, defect locations are unknown in advance, making it difficult to evaluate whether the agent's reports are correct. To address this, we control the level of guidance toward defects and propose two synthesis strategies.

\noindent\textbf{Defect-Oriented Task.} We manually collect reproduction trajectories that directly guide the agent to trigger defects, and synthesize action-level tasks from these trajectories. By minimizing exploration uncertainty, this strategy isolates the evaluation of defect recognition and reporting capabilities.

\begin{figure}[tp]
    \centering
    \includegraphics[width=1.0\linewidth]{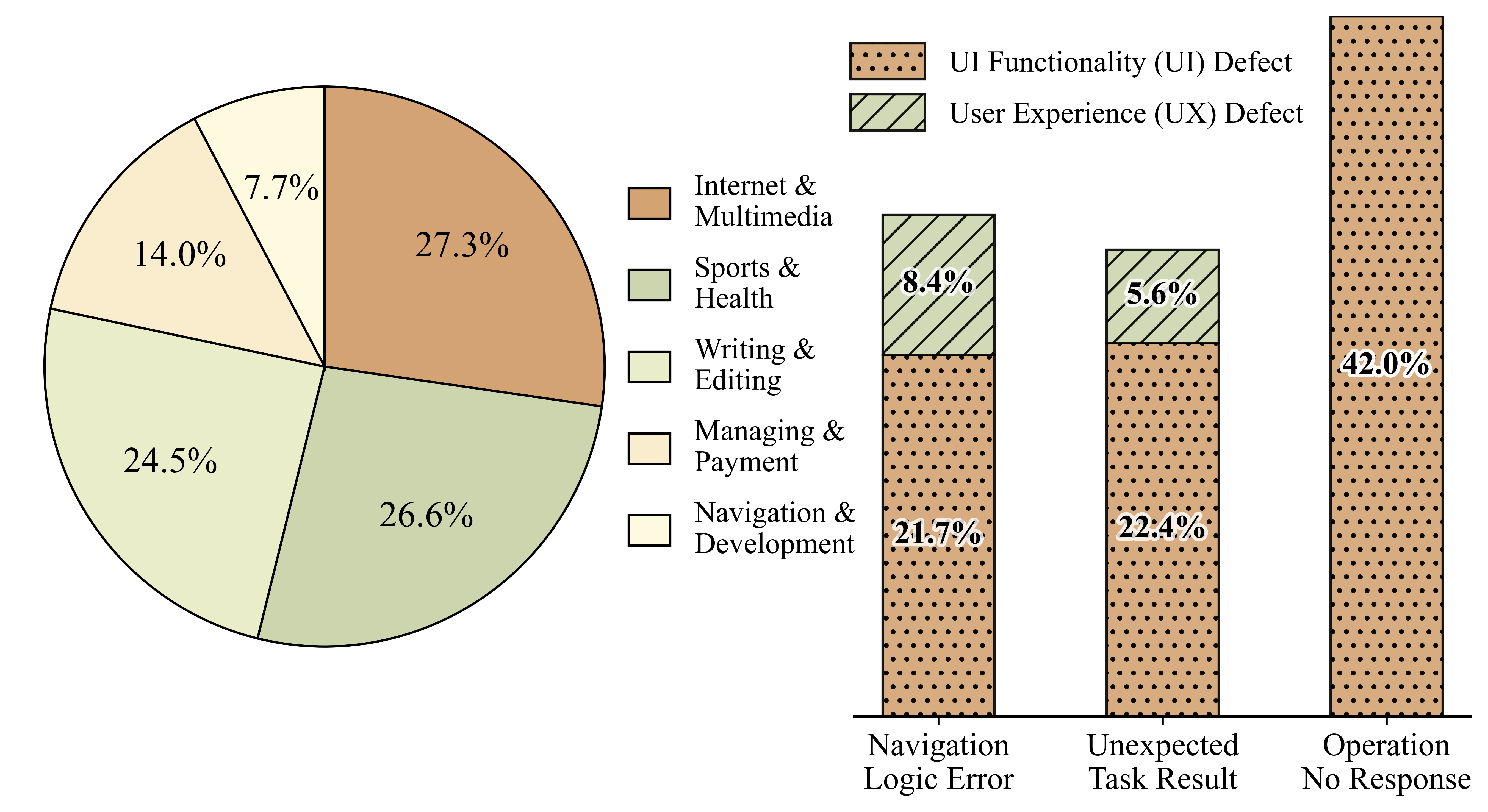}
    \caption{Defects distribution of GUITestBench}
    \label{fig:chart}
\end{figure}

\noindent\textbf{Exploration-Oriented Task.} We synthesize intent-level tasks where completing the task inevitably passes through the defect location. As shown in Figure~\ref{fig:guitestbench_case}, the synthesized task requires the agent to explore multiple cards. If the agent navigates correctly, it will encounter the target defect during exploration. By preserving exploration uncertainty, this strategy aims to evaluate the agent's end-to-end defect discovery capability, including navigation, recognition, and reporting.

\begin{figure*}[tp]
    \centering
    \includegraphics[width=1.0\linewidth]{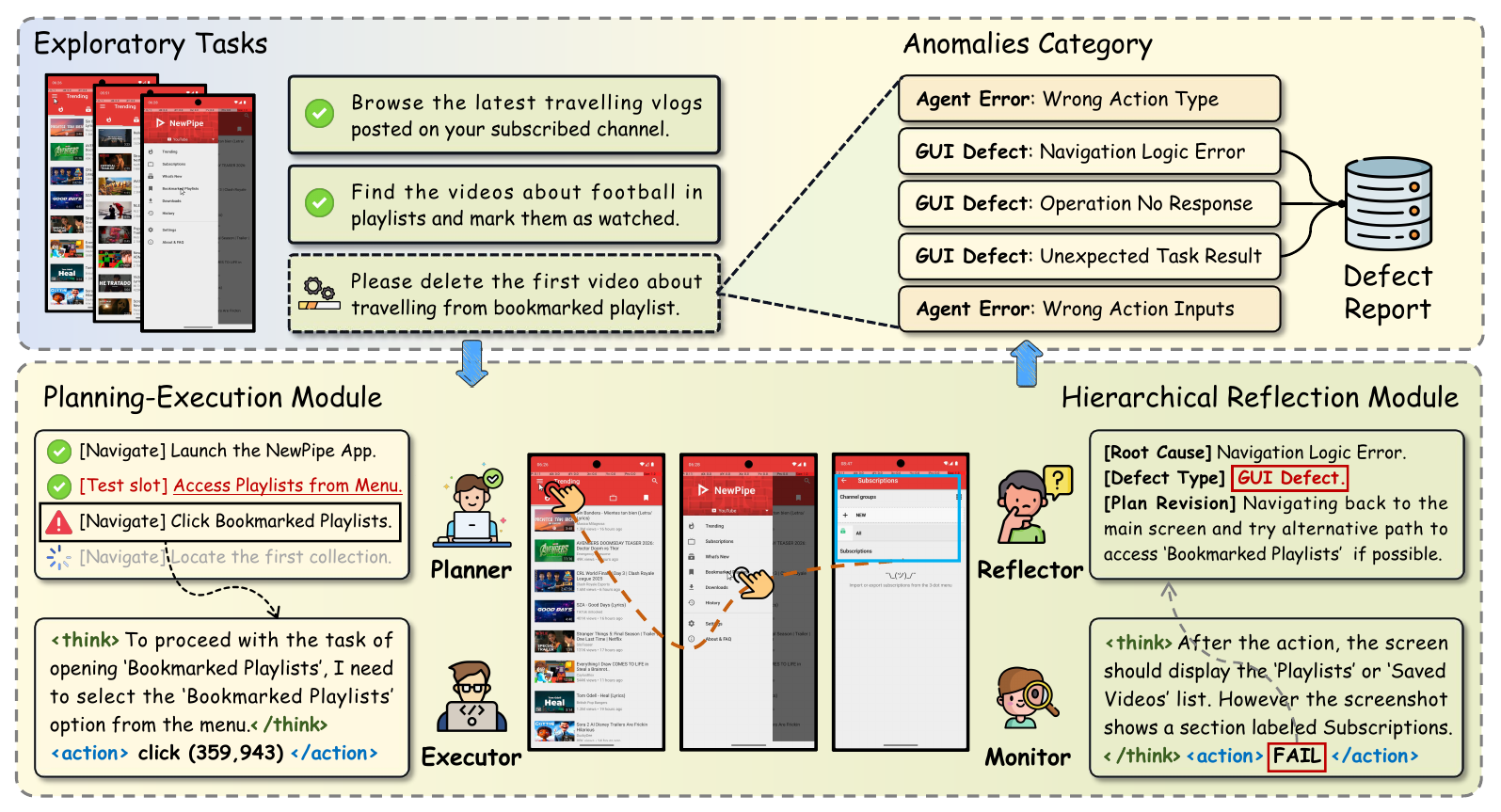}
    \caption{GUITester architecture. The system consists of four agents: (1) a Planner Agent for subtask planning and test intent generation; (2) an Executor Agent for GUI operation; (3) a Monitor Agent for capturing anomalies and controlling the execution process; and (4) a Reflector Agent for anomalies attribution and providing further adjustment for planning suggestions. The workflow are shown in Appendix~\ref{appsec:algorithm}.}
    \label{fig:guitester}
\end{figure*}

\subsubsection{Dataset Statistics}\label{GTBench_data_statistics}
We collect 26 defect from 12 applications across 5 diverse domains. Using the exploratory task synthesis strategies described above, we expand these scenarios into 143 navigation tasks. The detailed distribution across defect types and application domains is shown in Figure~\ref{fig:chart}. Based on defect-triggering mechanisms, defects fall into two categories: single-action defects, which are triggered by one action on a specific state (62.24\%), and multi-action defects, which require a sequence of prerequisite actions (37.76\%).

\subsection{Benchmark Evaluation}\label{GTBench_evaluation}
\subsubsection{Defect Detection Verification}\label{GTBench_defec_judgement}
To evaluate whether the agent successfully triggers a defect, we verify the exploration trajectory against the target defect specifications. Single-action defects have deterministic triggering conditions that can be precisely matched, while multi-action defects involve complex interaction sequences requiring flexible assessment. We thus employ two evaluation approaches:

\noindent\textbf{Rule-based Evaluation.} For single-action defects, we verify two conditions: (1) State matching: whether the agent successfully navigates to the screen where the defect resides; (2) Action matching: whether the agent executes the exact defect-triggering action.

\noindent\textbf{Judge-Model Evaluation.} For multi-action defects, we employ an LLM as the judge and provide it with detailed defect specifications, as shown in Figure~\ref{fig:guitestbench_case}, including preconditions, expected results, and screenshots before and after the triggering action. Given the agent's execution trajectory, the judge model determines whether the defect has been successfully triggered. The system prompt is shown in Appendix~\ref{appsec:multi_action_defects}.

\subsubsection{Evaluation Metrics}\label{GTBench_evaluation_metric}
Based on the above evaluation, we use Recall and F1 score to quantify the overall performance:

\noindent\textbf{Recall} measures the proportion of tasks in which the agent successfully identifies the target defect. Since each task corresponds to exactly one issued defect, we define: $\text{Recall}=|\mathcal{T}_{detected}|/|\mathcal{T}_{total}|$, where $\mathcal{T}_{detected}$ denotes the set of tasks correctly detecting the defect, and $\mathcal{T}_{total}$ is the set of all tasks.

\noindent\textbf{Precision} measures the proportion of tasks that correctly detect defects among all tasks where the agent reports GUI defects: $\text{Precision}=|\mathcal{T}_{detected}|/|\mathcal{T}_{declared}|$, where $\mathcal{T}_{declared}$ denotes the set of tasks in which the agent reported GUI defect.

\noindent\textbf{F1} is the harmonic mean of Precision and Recall:$$\text{F1} = \frac{2 \times \text{Precision} \times \text{Recall}}{\text{Precision} + \text{Recall}}$$which provides a balanced measure of the agent's defect discovery capability.

\section{GUITester}
We propose GUITester, a multi-agent framework that enables GUI agents to exploratory testing. As shown in Figure~\ref{fig:guitester}, GUITester comprises two core modules: (1) the Planning Execution Module (PEM, §\ref{GUITester_PEM}), which decomposes a navigation task into subtasks with embedded testing intents, guiding the agent to probe potential boundary behaviors; and (2) the Hierarchical Reflection Module (HRM, §\ref{GUITester_HRM}), which separates anomaly capture from attribution, ensuring defects are neither bypassed nor misattributed.

\subsection{Planning Execution Module}\label{GUITester_PEM}

\subsubsection{Planner Agent}

The Planner decomposes the navigation goal $g$ into a sequence of executable subtasks: $$\{s_1, s_2, \dots, s_n\} = \text{Planner}(g, o, h)$$where $o$ denotes the current observation and $h$ represents the historical context. Each subtask $s_i$ is either a navigation subtask that advances toward the goal, or a test intent that probes potential defects.

\noindent\textbf{Test Intent Generation}: Agent navigation tends to follow the shortest path to complete tasks, potentially missing defects hidden in specific interaction contexts. To increase the defect exposure, the Planner embeds test intents that guide the agent to explore boundary behaviors during navigation. We design three patterns based on how defects manifest in GUI applications, as shown in Appendix~\ref{app-test_intent}. These test intents are interleaved with navigation subtasks without disrupting task completion.

\subsubsection{Executor Agent}
The Executor translates subtasks into executable actions. Given a subtask $s_i$ from the Planner, the Executor observes the current environment state $o_t$ and generates an action:
$$a_t = \text{Executor}(s_i, o_t, \{a_{t'}\}_{t'<t})$$
where $\{a_{t'}\}_{t'<t}$ denotes the action history within the current subtask. We adopt existing GUI agent models (e.g., UI-TARS~\cite{UI-TARS}, GUI-Owl~\cite{mobile-agent-v3}) as the Executor, which generates actions through chain-of-thought reasoning.

\input{table-overall}

\subsection{Hierarchical Reflection Module}\label{GUITester_HRM}

\subsubsection{Monitor Agent}

As mentioned above, GUI agents may bypass execution anomalies by exploring alternative paths or waiting for user feedback, potentially overlooking genuine GUI defects. To address this, the Monitor observes the environment's response after each action and determines the execution state:
$$c_t = \text{Monitor}(s_i, o_t, a_t, o_{t+1})$$
where $c_t \in \{\texttt{DONE}, \texttt{FAIL}, \texttt{CONTINUE}\}$. Specifically, the Monitor focuses solely on capturing whether an anomaly occurs without attributing its cause. When anomalies such as unresponsive operations or unexpected state transitions are captured, it terminates the subtask and issues a \texttt{FAIL} state, preventing the agent from bypassing potential defects. In contrast, the \texttt{CONTINUE} state allows the Executor to proceed with the current subtask, while \texttt{DONE} signals the Planner to advance to the next subtask.

\subsubsection{Reflector Agent}
The Reflector is responsible for attributing anomalies captured by the Monitor. When a \texttt{FAIL} state is received, the Reflector analyzes the execution trajectory to distinguish between agent navigation errors and genuine GUI defects:
$$r = \text{Reflector}(s_i, \tau, o_t)$$
where $\tau = \{(o_{t'}, a_{t'})\}_{t'=1}^{t}$ represents the execution trajectory, and $r$ denotes the attribution result.

\noindent\textbf{Visual Attribution.} To enable accurate attribution, we visualize the Executor's actions on corresponding screenshots by marking interaction points~\cite{GUI-Actor,OmniParser}. This allows the Reflector to clearly identify whether the anomaly stems from ineffective operations (e.g., misaligned click) or genuine defects (e.g., unresponsive buttons).

\noindent\textbf{Reflection Feedback.} After attribution, the Reflector provides feedback to the Planner to prevent repeated failures at the same location. For navigation subtasks attributed to agent errors, the Reflector generates corrective suggestions to guide subsequent planning. For failed test intent subtasks, we prevent them from affecting  planning regardless of the attribution result, allowing the agent to continue exploration from the current state.

\section{Experiments}
We conduct comprehensive experiments to answer two research questions:
\begin{itemize}[leftmargin=1em,itemsep=-2pt]
    \item \textbf{RQ1}: How do existing GUI agents perform in exploratory GUI testing scenario?
    \item \textbf{RQ2}: How effective is GUITester in addressing the challenges of exploratory GUI testing?
\end{itemize}
To answer these questions, we design quantitative evaluations on GUITestBench (§\ref{Exp_setup}, §\ref{Exp_results}) and further validate GUITester's practical effectiveness through case studies on released apps (§\ref{Exp_cases}).

\subsection{Experiment Setup}\label{Exp_setup}
\noindent\textbf{Baseline GUI Agents.} We select GUI agents with visual grounding capabilities as our evaluation baselines. These agents can interpret interface states from screenshots and generate corresponding interaction actions. To enable defect reporting during navigation, we augment the action instructions with explicit testing intent through prompt wrapping (see Appendix~\ref{appsec:GUIAgents_setup} for details). We evaluate six open-source GUI agent models from three families: MAI-UI (8B)~\cite{MAIUI}, GUI-Owl (7B/32B)~\cite{mobile-agent-v3}, and UI-TARS (7B/72B/1.5-7B)~\cite{UI-TARS}. Additionally, we evaluate Mobile-Agent-V3~\cite{mobile-agent-v3}, a multi-agent workflow designed for navigation tasks, powered by GUI-Owl-32B.

\noindent\textbf{GUITester Setup.} GUITester comprises four collaborative agents. The \textbf{Planner} handles task decomposition and test intent generation, powered by \texttt{Qwen3-VL-Plus}. The \textbf{Executor} performs GUI actions with a low sampling temperature (0.1) to ensure behavioral stability; we evaluate UI-TARS-72B, UI-TARS-1.5-7B, and GUI-Owl-32B as Executor backbones, respectively. The action space is determined by the Executor model (see Appendix~\ref{appsec:action-space} for detailed action space). The \textbf{Monitor} detects anomalies using \texttt{GPT-4o}. The \textbf{Reflector} performs defect attribution through trajectory analysis, also powered by \texttt{Qwen3-VL-Plus}. All experiments are conducted on Android emulators with 1080×2400 resolution.

\input{table-analysis}

\noindent\textbf{Evaluation Setup.} All models are evaluated on the 143 tasks of GUITestBench with three independent runs. Pass@1 results are computed by averaging the scores across the runs, while Pass@3 indicates whether at least one successful detection occurs. We employ \texttt{Claude-4-Sonnet} as the Judge.

\subsection{Experimental Results}\label{Exp_results}

\subsubsection{RQ1}
\noindent\textbf{Overall Performance.} As shown in Table~\ref{tab:main_results}, all baseline agents achieve F1 below 25\% under the Pass@1 setting, with the best-performing model, UI-TARS-1.5-7B, reaching only 22.95\%. Under the Pass@3 setting, this improves to 33.35\%, yet over 70\% of defects remain undetected. Since Pass@3 reflects the upper bound of detection capability across multiple attempts, these results indicate that exploratory GUI defect discovery remains a challenging task for existing GUI agents. Appendix~\ref{appsec:other_failure} provides detaild analysis of failure cases.

\noindent\textbf{UI Defects vs. UX Defects.} All baseline agents demonstrate a certain degree of detection capability for UI defects, whereas their performance on UX defects is near zero, with only UI-TARS-1.5-7B achieving non-zero results (F1 of 12.95\% on UX-UTR and 18.38\% on UX-NLE). This disparity arises from the inherent differences between the two defect categories: UI defects typically manifest as immediate visual anomalies that can be identified through single-frame analysis, while UX defects cannot be attributed to any specific operation and require the agent to comprehend the entire interaction sequence to identify defects. Although existing agents incorporate historical screenshots during inference, their training objectives focus on action prediction, leveraging past information solely to determine the next action, rather than to retrospectively assess the interaction logic.

\noindent\textbf{Analysis Across Defect Types.} Across all three defect types (ONR, UTR, and NLE), the baseline agents exhibit consistently poor performance. Despite their distinct manifestations, these defect types share a common requirement: the ability to perceive discrepancies between expected and actual states and to correctly attribute their causes. However, current agents fail to meet this requirement due to task-success bias inherited from their training objectives, which prioritize generating correct actions over verifying environment states. This leads to two failure modes: (1) agents passively accept defective states as normal and continue toward task completion; or (2) when agents do perceive anomalies, they preferentially attribute failures to their own operational errors rather than questioning the GUI itself, causing genuine GUI defects to remain undiscovered.

\vspace{0.15cm}
\begin{dialoguebox}{agentcolor}{ANSWER TO RQ1}
Existing GUI agents struggle with exploratory defect discovery (Recall below 20\% at Pass@1). We identify two limitations: (1) the lack of ability to retrospectively analyze interaction logic, and (2) the lack of state verification due to task-success bias. These limitations cause agents to either passively accept defective states or misattribute anomalies to their own execution errors.
\end{dialoguebox}

\begin{figure*}[tp]
    \centering
    \includegraphics[width=0.98\linewidth]{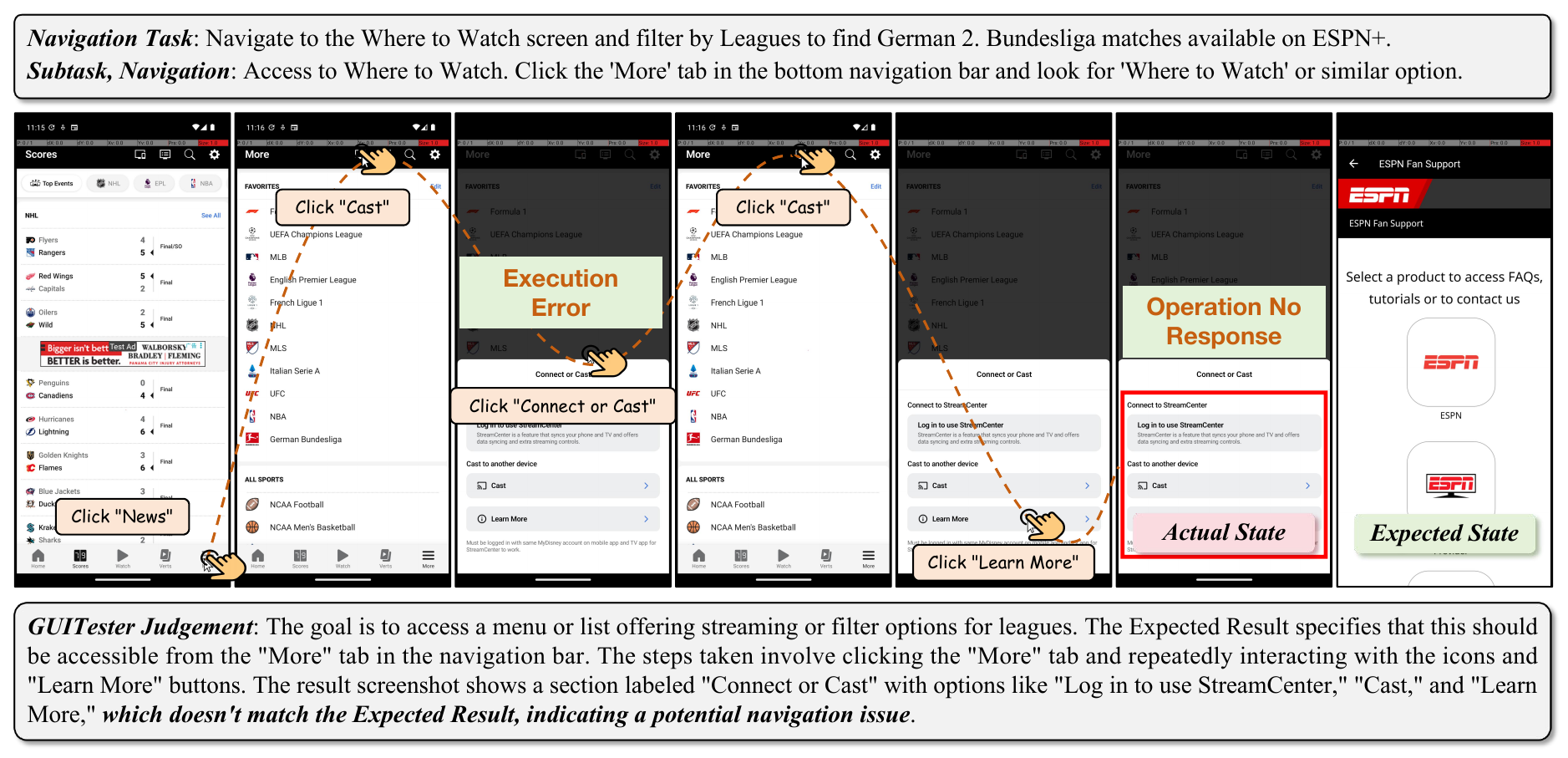}
    \caption{A defect detected by GUITester on ESPN (v8.6.0, November 2025). The "Learn More" button in the Cast panel is visually present but functionally non-responsive, failing to navigate to the expected support page.}
    \label{fig:final-case}
\end{figure*}

\subsubsection{RQ2}
\noindent\textbf{Overall Performance}. As shown in Table~\ref{tab:main_results}, under the Pass@3 setting, GUITester achieves significant improvements. For example, with UI-TARS-72B, the overall F1 increases from 19.55\% to 43.70\%, demonstrating that GUITester can effectively enhance the defect detection capability of GUI agents in exploratory GUI testing.

\noindent\textbf{Analysis Across Task Types.} As shown in Table~\ref{tab:main_results-2}, although baseline agents exhibit some detection capability in the Defect-Oriented setting, their performance drops substantially in the Exploration-Oriented setting, indicating that their defect discovery capability is difficult to leverage during autonomous exploration. In contrast, GUITester maintains relatively consistent performance across both settings. We attribute this improvement to PEM's test intent embedding, which actively drives the Executor to probe potential boundary behaviors rather than passively waiting to encounter defects, thereby increasing defect exposure during exploration (see Appendix~\ref{fig:guitester-fox_news}, where PEM's boundary testing on a search field uncovered a defect).

\noindent\textbf{Analysis on Defect Complexity.} As shown in Table~\ref{tab:main_results-2}, GUITester achieves substantial improvements on Single-Action defects. Single-Action defects manifest immediately after a single operation, making them detectable through per-step state verification. HRM is well-suited for this scenario: the Monitor captures state anomalies right after each action, and the Reflector attributes them before the agent proceeds, ensuring that transient defects are not overlooked or misattributed. On Multi-Action defects, GUITester also outperforms baselines, though with relatively smaller gains. Triggering such defects requires not only anomaly detection and attribution, but also accurate navigation to specific action sequences, placing higher demands on planning and execution capabilities.

\vspace{0.15cm}
\begin{dialoguebox}{agentcolor}{ANSWER TO RQ2}
GUITester significantly improves detection performance (overall F1 reaching 48.90\%). Our results suggest: (1) proactive exploration through embedded test intents, rather than passively waiting to encounter defects; and (2) decoupling anomaly detection from attribution to enable active state verification and defect identification.
\end{dialoguebox}

\subsection{Case Studies on Released APPs}\label{Exp_cases}
We deploy GUITester on publicly released historical versions of real-world applications to validate its practical effectiveness. Figure~\ref{fig:final-case} shows an ONR defect discovered on ESPN app during exploratory testing. Within the same trajectory, the agent encounters two anomalies: clicking outside the "Connect or Cast" popup causing it to dismiss unexpectedly, and a non-responsive "Learn More" button. HRM correctly attributes the former to agent error and initiates self-correction, while identifying the latter as a GUI defect. This demonstrates that GUITester can accurately attribute anomalies, which is a critical capability for reducing false positives in real-world testing scenarios. More GUI defect detection results are available in Appendix~\ref{appsec:more-cases}.

\section{Conclusion}
This paper identifies two key challenges preventing GUI agents from effective exploratory testing: Goal-Oriented Masking and Execution-Bias Attribution. We introduce GUITestBench, the first interactive benchmark for evaluating defect discovery capabilities, and propose GUITester, a multi-agent framework that proactively probes boundary behaviors and decouples anomaly detection from attribution. Experiments demonstrate that GUITester enables effective defect exposure and accurate defect reporting, validating the feasibility of autonomous exploratory GUI testing and opening up a new direction for GUI agent-based quality assurance.

\section{Limitations}
While applying GUI agents to exploratory GUI testing opens promising avenues, we find some failure cases caused by practical challenges that warrant further investigation, summarized as follows. Detailed analysis is provided in Appendix~\ref{appsec:failure}.

\noindent(1) \textbf{The Wait-or-Miss Dilemma.} Real-world environments are noisy (e.g., network fluctuations, server lag, and page timeouts). We observed GUI agents occasionally reporting slow-loading pages (6-7 seconds) as defects. The naive fix of "just wait longer" creates its own problems: testing efficiency plummets, and worse, some genuine defects manifest as fleeting millisecond glitches that extended waiting would miss entirely. Distinguishing environmental delays from true anomalies remains an open challenge in real-world deployment.

\noindent(2) \textbf{Monitor Capability Boundaries.} Accurate defect detection requires the Monitor to predict expected states after each action. However, the Monitor's effectiveness is bounded by its domain knowledge and assumptions about application behavior. For instance, it may misjudge valid navigation paths due to unfamiliarity with domain-specific relationships (e.g., UFC as a subsidiary of MMA). Integrating application-specific knowledge or more capable vision-language models could improve precision.

\noindent(3) \textbf{Small Action Space, Narrow Testing Scope.} Current GUI agents are equipped with action spaces designed for autonomous navigation. However, real-world testing demands richer interactions. Consider a trading application: users routinely \texttt{zoom in/out} on stock price charts to examine minute-by-minute fluctuations, yet no existing GUI agent can perform this gesture. Such capability gaps leave some defects unexplored.

\noindent(4) \textbf{Broader Defect Coverage.} This work focuses on interactive defects. However, GUIs can fail in other ways: overlapping elements, misaligned text, truncated labels. Expanding defect coverage to include such layout issues would enable more comprehensive quality assurance.

\vspace{0.1cm}
\noindent Future work will focus on these directions: (1) developing robust strategies to handle real-world environmental noise, (2) enhancing HRM's domain understanding, (3) enriching action spaces for broader testing scenarios, and (4) expanding defect coverage for comprehensive GUI quality assurance.

\bibliography{custom}


\appendix
\section{GUITestBench}
\subsection{Exploration Task Synthesis}\label{appsec:instruction_synthesis}

This section details the synthesis procedures for the Defect-Oriented and Exploration-Oriented tasks.

\noindent\textbf{Defect-Oriented Task.} Given a manually collected reproduction trajectory that reaches the defect, we employ an LLM to abstract the action sequence into a natural language instruction. The LLM is provided with the application's functional context and the verified interaction sequence, producing a goal-directed instruction that guides the agent directly toward the defect location.

\noindent\textbf{Exploration-Oriented.}
To construct tasks where the defect serves as a necessary waypoint, we adopt a three-stage synthesis-and-filter pipeline:
\begin{itemize}[leftmargin=1em,itemsep=-2pt]
    \item \textit{Pre-defect Intent Synthesis}. Using the reproduction trajectory from the initial state to the defect, we prompt an LLM to generate multiple navigation intents that would lead to the defect location.
    \item \textit{Post-defect Intent Synthesis}. Starting from the defect page, we prompt an LLM to generate plausible continuation intents, which may navigate deeper into the app, return to previous screens, or explore sibling functionalities.
    \item \textit{Combination and Filtering}. We combine pre-defect intents, the defect-triggering actions, and post-defect intents into composite task instructions. For instance, 5 pre-defect intents and 3 post-defect intents yield 15 candidate tasks. Each candidate is then executed by a GUI agent, and only those where the agent successfully reaches the defect location are retained.
\end{itemize}
\noindent This pipeline ensures that the resulting tasks possess a bottleneck structure: completing the task necessitates traversing the defect state, enabling evaluation of end-to-end defect discovery under realistic exploration scenarios.

\subsection{Multi-Action Defect Verification}\label{appsec:multi_action_defects}
Multi-action defects require specific sequences of operations to trigger, making automated verification more challenging than single-action defects. We employ an LLM-based judge to determine whether the agent's trajectory successfully reproduces the target defect. As shown in Table~\ref{prompt:judge_model}, the judge model receives four inputs: (1) the defect description specifying preconditions, trigger actions, and expected results; (2) reference screenshots demonstrating the defect behavior; (3) the agent's execution trajectory; and (4) screenshots of the agent's final state. The judge then performs a three-step verification: checking whether preconditions are satisfied, whether the trigger action is correctly executed, and whether the final state matches the expected defect behavior. We highlight two key aspects in the prompt design: (1) Strict action sequence matching: A trajectory that only satisfies preconditions but misses the trigger action is marked as failure, since the trigger action is essential for defect manifestation. (2) Flexible input values: The agent may use different input values from the defect description examples, as long as the action sequence and interaction pattern remain consistent.

\subsection{More Examples of GUITestBench}\label{appsec:GUITestBench_cases}
\begin{figure*}[!t]
    \centering
    \includegraphics[width=1.0\linewidth]{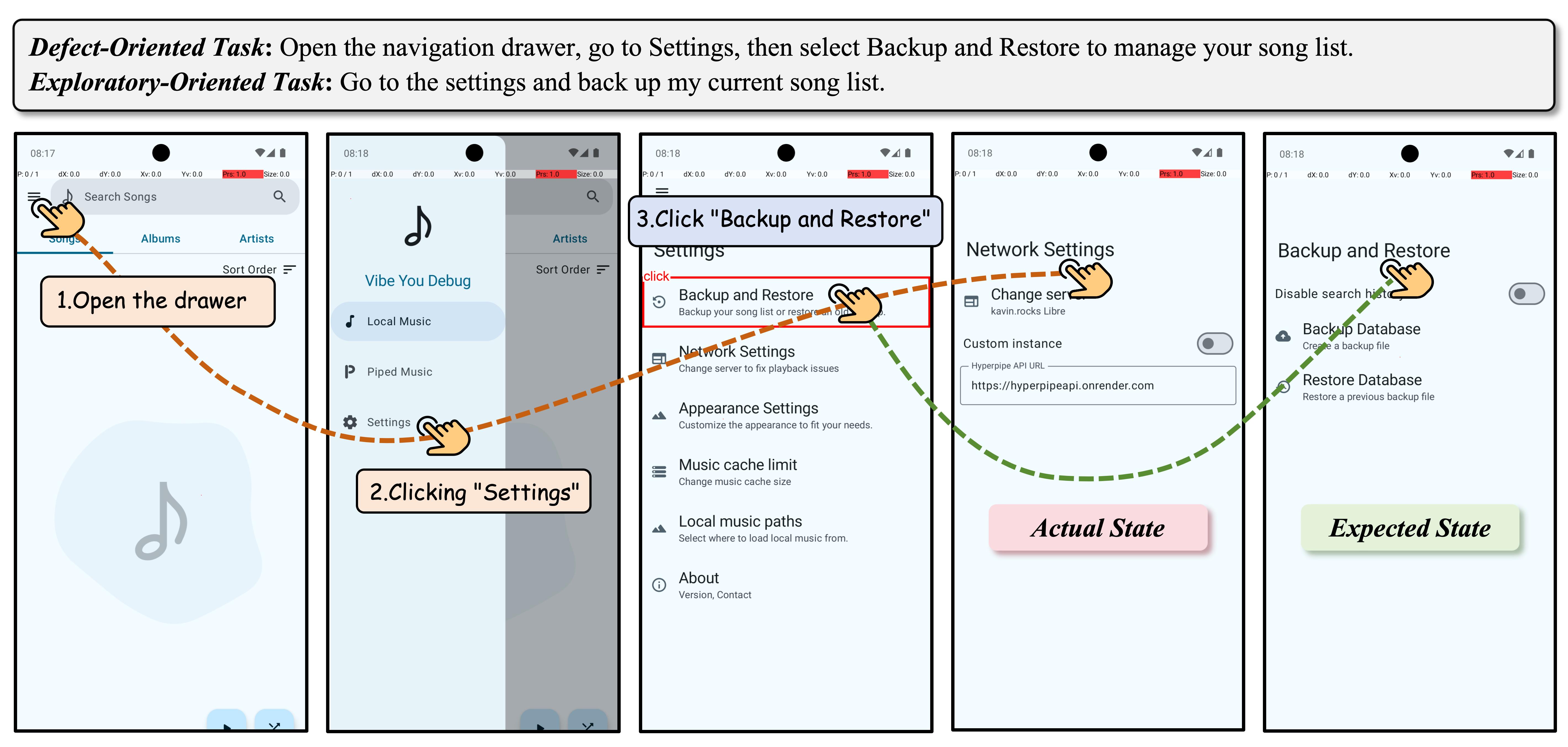}
    \caption{Example of UI-NLE (Navigation Logic Error) defect. The task requires navigating to "Backup and Restore" in Settings. After clicking "Backup and Restore", the app incorrectly navigates to "Network Settings" instead of the expected "Backup and Restore" page, demonstrating a navigation logic error where the destination does not match the triggered element.}
    \label{fig:guitestbench-02}
\end{figure*}

\begin{figure*}[tp]
    \centering
    \includegraphics[width=1.0\linewidth]{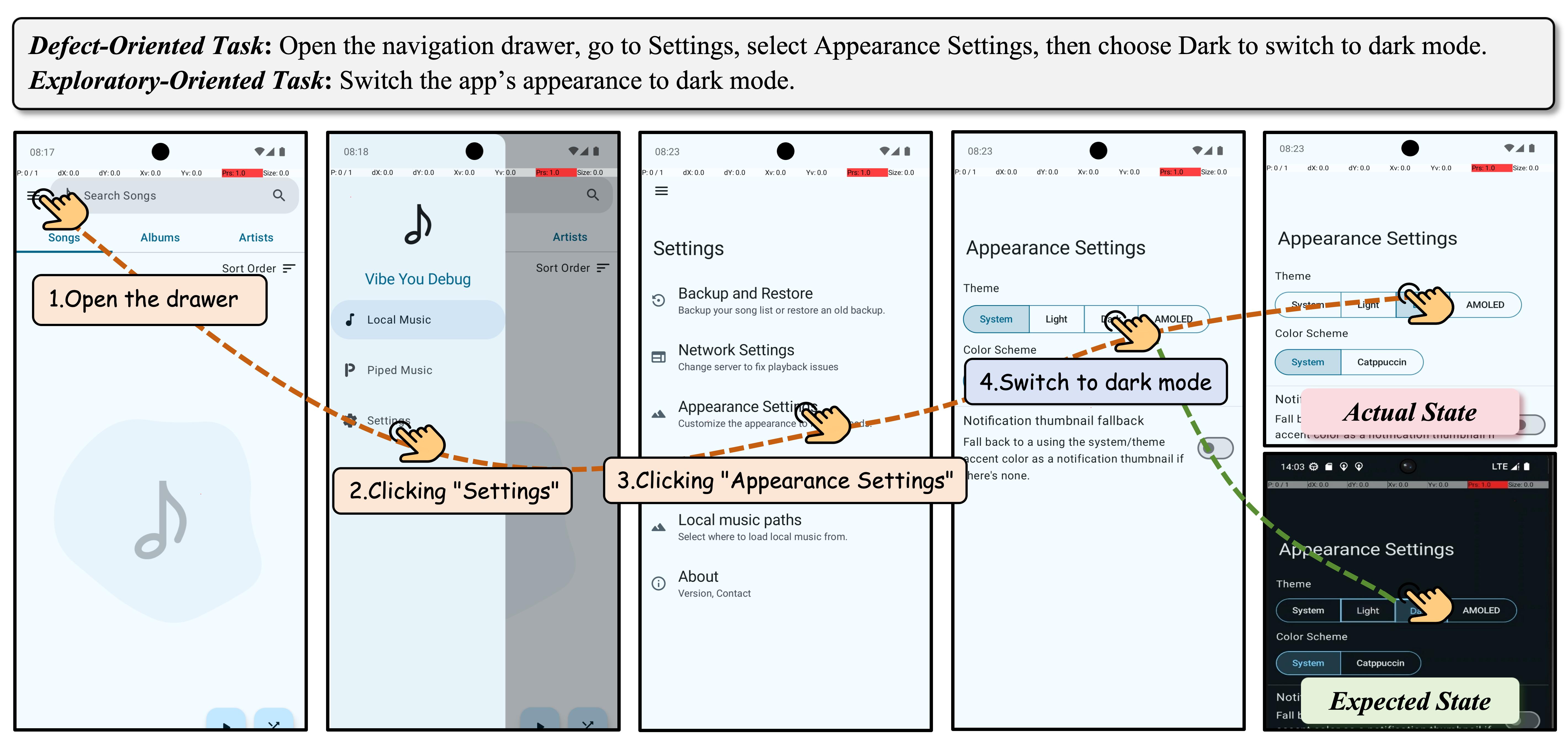}
    \caption{Example of UI-ONR (Operation No Response) defect. The task requires switching the app's appearance to dark mode. After selecting "Dark" in Appearance Settings, the desired appearance does not appear, indicating that the dark mode switch operation did not produce the expected response and the interface state remains inconsistent with user intent.}
    \label{fig:guitestbench-03}
\end{figure*}

\begin{figure*}[tp]
    \centering
    \includegraphics[width=1.0\linewidth]{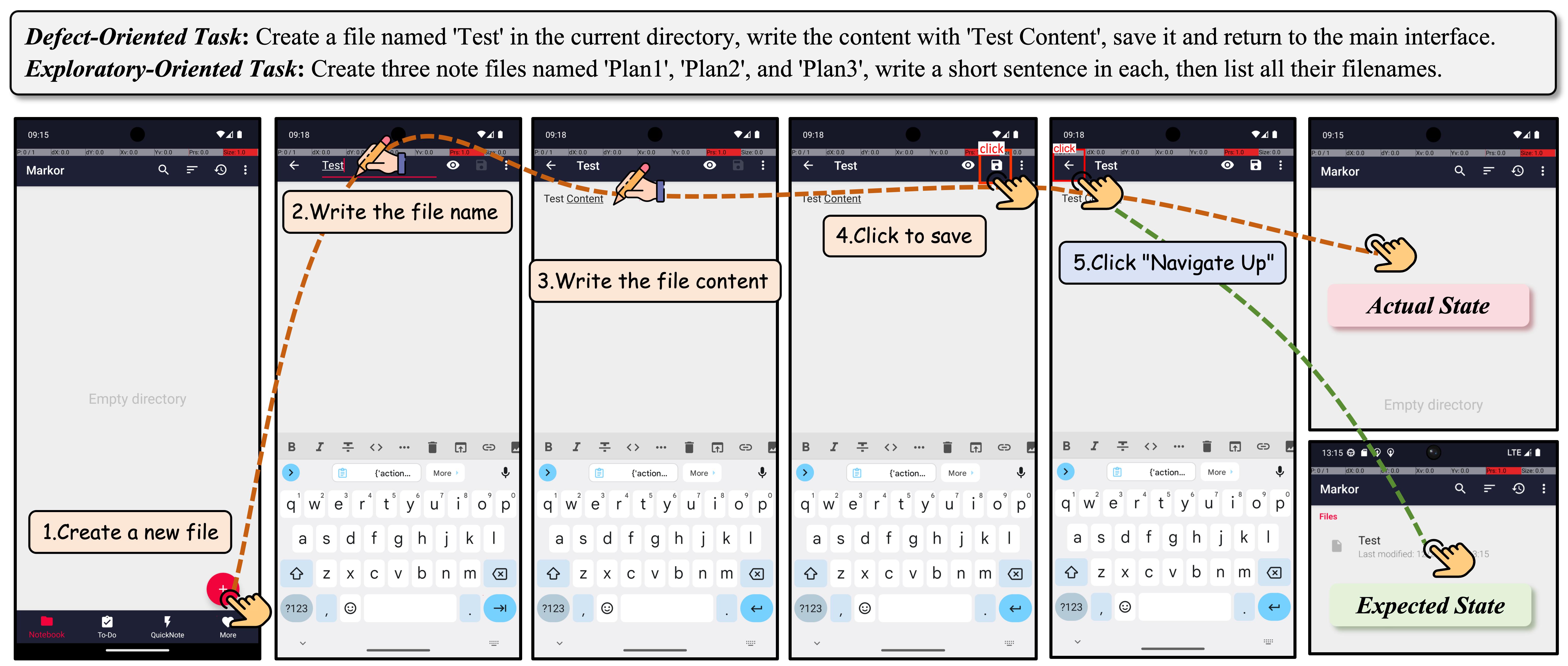}
    \caption{Example of UX-UTR (User Experience, Unexpected Task Result) defect. The task requires creating a file named "Test", writing content, saving it, and returning to the main interface. Although each individual operation succeeds, the final state shows an empty directory without the created file, indicating that the overall task result does not match user expectations despite seemingly correct step-by-step execution.}
    \label{fig:guitestbench-04}
\end{figure*}

\begin{figure*}[tp]
    \centering
    \includegraphics[width=1.0\linewidth]{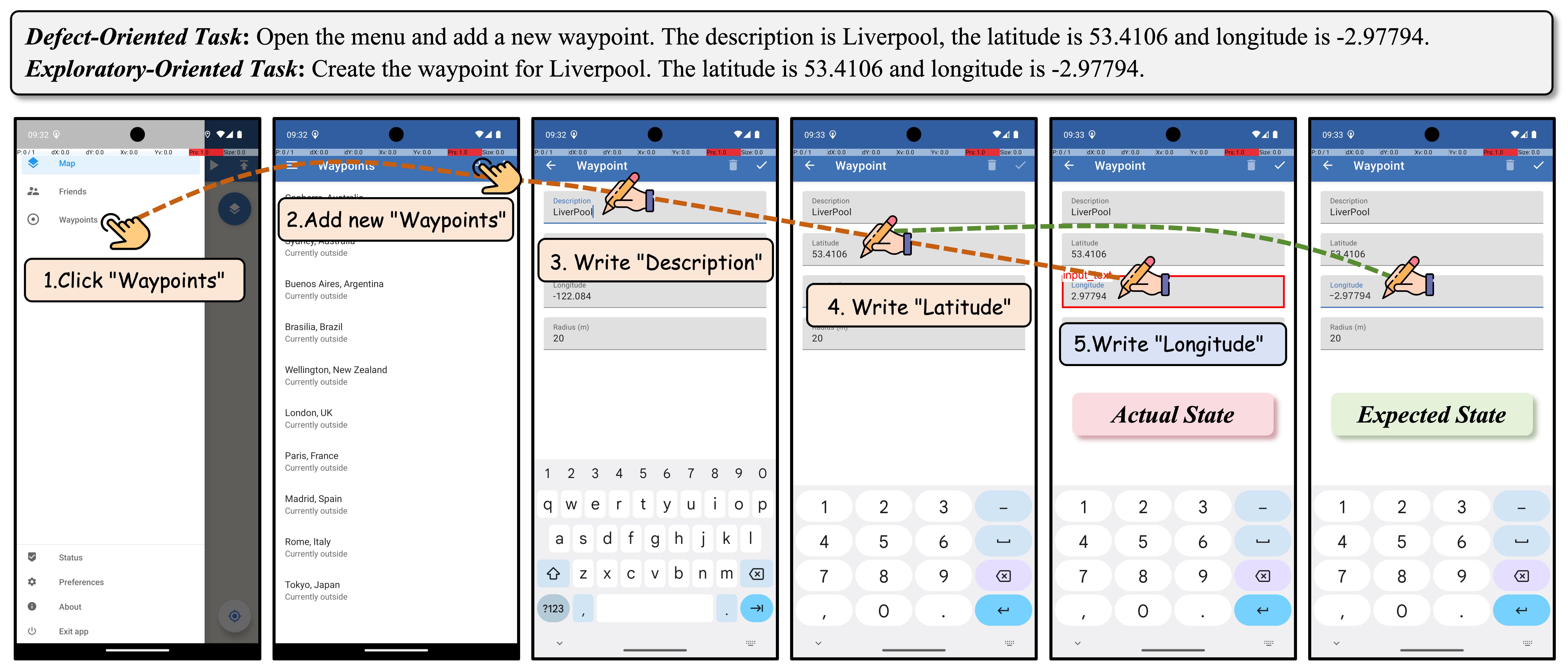}
    \caption{Example of UI-UTR (Unexpected Task Result) defect. The task requires adding a waypoint for Liverpool with specific coordinates (latitude: 53.4106, longitude: -2.97794). After entering all information, the longitude value is incorrectly saved as "2.97794" instead of "-2.97794", demonstrating an unexpected task result where the input data is not correctly preserved.}
    \label{fig:guitestbench-05}
\end{figure*}

We provide representative examples for each defect category in GUITestBench. Each example illustrates both the Defect-Oriented task (with step-by-step guidance) and the Exploration-Oriented task (with only high-level intent), along with the reproduction trajectory and the contrast between actual and expected states. Figure \ref{fig:guitestbench-02} shows a Navigation Logic Error (UI-NLE) where clicking an element leads to an incorrect destination. Figure \ref{fig:guitestbench-03} demonstrates an Operation No Response (UI-ONR) defect where the interface fails to respond as expected. Figure \ref{fig:guitestbench-04} presents a User Experience defect (UX-UTR) where individual operations succeed but the overall task outcome is incorrect. Figure \ref{fig:guitestbench-05} illustrates an Unexpected Task Result (UI-UTR) where user input is not correctly preserved.

\begin{algorithm}[!t]
    \footnotesize
    \setlength{\algomargin}{0.5em}  
    \begin{minipage}{\linewidth}
    \caption{GUITester Workflow}
    \SetAlgoVlined
    \SetInd{0.3em}{0.6em}  
    
    \label{alg:guitester_final}
    
    \KwIn{
        navigation goal $g$, 
        current observation $o_t$, 
        history $h$,
        $\texttt{MaxSteps}$ \tcp*[r]{\textcolor{gray!30}{max trajectory steps}}
    }
    \KwOut{
        $a_t$ or $\texttt{NOOP}$ \tcp*[r]{\textcolor{gray!30}{action, or terminate state}}
    }
    \LinesNotNumbered
    \SetKwInput{KwInit}{Init}
    \KwInit{
        $\tau \leftarrow \emptyset$ \tcp*[r]{\textcolor{gray!30}{trajectory for the current subtask}} \\ \texttt{replan}$\leftarrow$\texttt{True}; \tcp*[r]{\textcolor{gray!30}{adjust the current plan}} \\
        \texttt{next\_subtask}$\leftarrow$\texttt{False}; \tcp*[r]{\textcolor{gray!30}{get the next subtask}} \\
        \texttt{check\_status}$\leftarrow$\texttt{False}; \tcp*[r]{\textcolor{gray!30}{monitoring the state}} \\
        \texttt{reflect}$\leftarrow$\texttt{False}; \tcp*[r]{\textcolor{gray!30}{reflect the failure}} \\
        \texttt{send\_action}$\leftarrow$\texttt{False}; \tcp*[r]{\textcolor{gray!30}{send action to env.}}
    }
    \LinesNumbered
    
    \BlankLine
    \texttt{SyncState}$(o_t)$ \tcp*[r]{\textcolor{gray!30}{share $o_t$ with all agents}}
    \While{$\lnot$\texttt{send\_action}}{
        \tcp{\textcolor{blue}{Planner: decompose task into subtasks}}
        \If{\texttt{replan}}{
            $\{s_1,\dots,s_n\} \leftarrow \text{Planner}(g,o_t,h,r)$\;
            \texttt{replan} $\leftarrow$ \texttt{False}, \texttt{next\_subtask} $\leftarrow$ \texttt{True}\;
        }
        
        \BlankLine
        \If{\texttt{next\_subtask}}{
            $s \leftarrow \text{Planner.GetLatestSubtask}()$\;
            \If{$s=\varnothing$}{\Return $\{\texttt{NOOP}\}$ \tcp*[r]{\textcolor{gray!30}{all tasks completed}}}
            \texttt{next\_subtask} $\leftarrow$ \texttt{False}\;
        }
        
        \BlankLine
        \tcp{\textcolor{blue}{Monitor: capture execution anomalies}}
        \If{\texttt{check\_status}}{
            $c_t \leftarrow \text{Monitor}(s,o_{t},\tau)$ \tcp*[r]{\textcolor{gray!30}{$\{\texttt{DONE},\texttt{FAIL},\texttt{CONTINUE}\}$}}
            \uIf{$c_t==\texttt{DONE}$}{
                $h \leftarrow h \cup \{(s,\texttt{DONE})\}$, $\tau \leftarrow \emptyset$\;
                \texttt{next\_subtask} $\leftarrow$ \texttt{True}, \texttt{check\_status} $\leftarrow$ \texttt{False}\;
            }
            \uElseIf{$c_t==\texttt{FAIL}$}{
                $h \leftarrow h \cup \{(s,\texttt{FAIL})\}$\;
                \texttt{reflect} $\leftarrow$ \texttt{True}, \texttt{check\_status} $\leftarrow$ \texttt{False}\;
            }
            \Else{
                \texttt{check\_status} $\leftarrow$ \texttt{False}\;
            }
        }
        
        \BlankLine
        \tcp{\textcolor{blue}{Reflector: attribute anomalies and recovery}}
        \If{\texttt{reflect}}{
            $r \leftarrow \text{Reflector}(s,o_{t},\tau,h)$ \tcp*[r]{\textcolor{gray!30}{analyze trajectory $\tau$}}
            
            \uIf{$r==\texttt{AGENT\_ERROR} \land |\tau|<\text{MaxSteps}$}{
                \texttt{reflect} $\leftarrow$ \texttt{False}\;\tcp*[r]{\textcolor{gray!30}{self-correct without replanning}}
            }
            \Else{
                $\tau \leftarrow \emptyset$ \tcp*[r]{\textcolor{gray!30}{reset trajectory}}
                \texttt{replan} $\leftarrow$ \texttt{True}, \texttt{reflect} $\leftarrow$ \texttt{False}\;
            }
        }
        
        \BlankLine
        \tcp{\textcolor{blue}{Executor: generate action for current subtask}}
        \If{$\neg\texttt{replan}\land \neg\texttt{next\_subtask} \land \neg\texttt{check\_status} \land \neg\texttt{reflect}$}{
            $a_{t+1} \leftarrow \text{Executor}(s,o_t,\tau)$; \quad $\tau \leftarrow \tau \cup \{(a_{t+1},o_{t})\}$\;
            \texttt{check\_status} $\leftarrow$ \texttt{True}\;
            \texttt{send\_action} $\leftarrow$ \texttt{True} \tcp*[r]{\textcolor{gray!30}{send action to env.}}
        }
    }
    \end{minipage}
\end{algorithm}

\begin{figure*}[!t]
    \centering
    \includegraphics[width=1.0\linewidth]{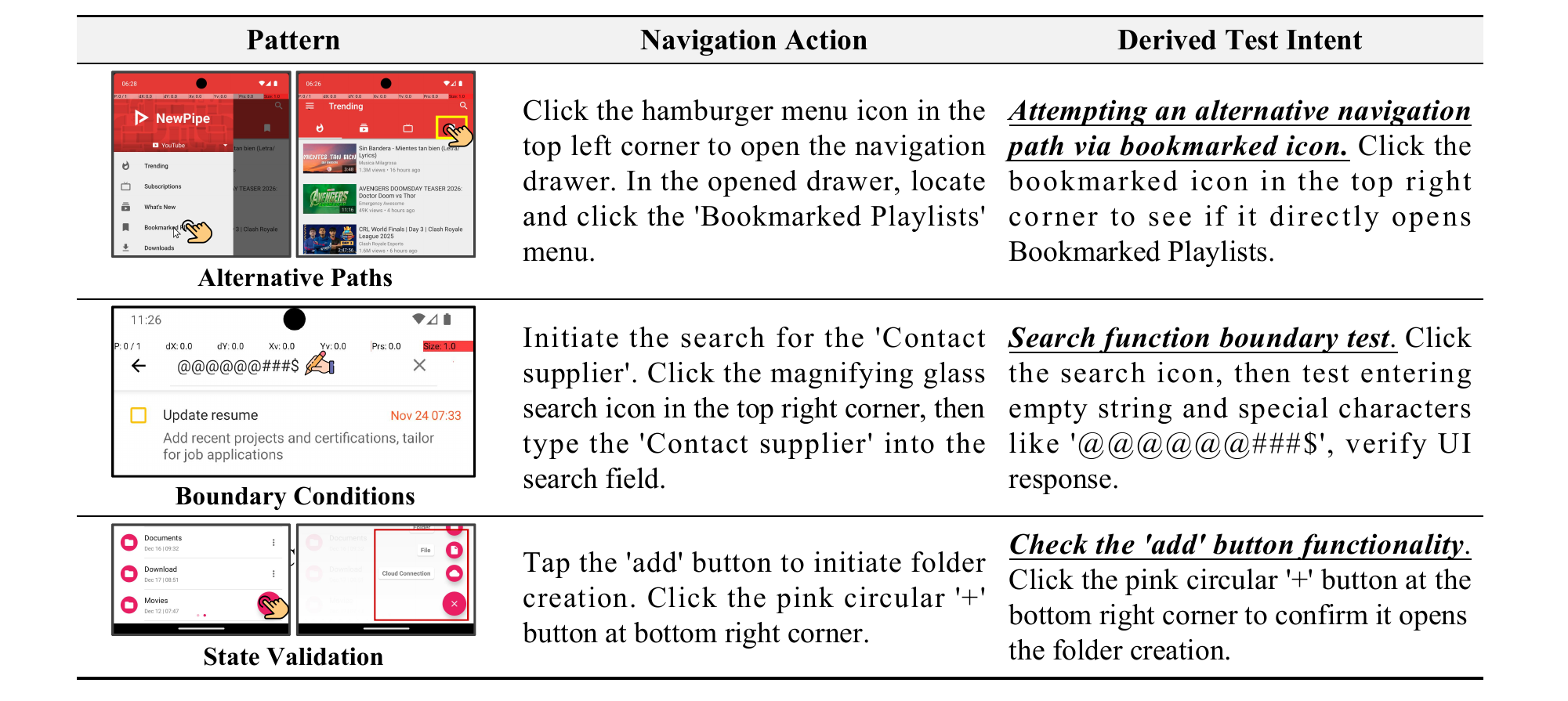}
    \caption{Three test intent patterns designed to increase defect exposure. For each pattern, we show an example navigation action (middle column) and its corresponding test intent (right column) that probes potential boundary behaviors. Alternative Paths explores different entry points to the same functionality; Boundary Conditions tests edge cases such as empty or special character inputs; State Validation verifies UI element responsiveness and state transitions.}
    \label{app-test_intent}
\end{figure*}

\section{GUITester}\label{appsec:algorithm}
Algorithm~\ref{alg:guitester_final} presents the complete workflow of GUITester. The system operates as a loop coordinated by four agents: \textbf{Planner}, \textbf{Executor}, \textbf{Monitor}, and \textbf{Reflector}.

\noindent\textbf{Planning.} The Planner decomposes the navigation task $g$ into a sequence of subtasks $\{s_1, \dots, s_n\}$, which includes both navigation subtasks and test intents (as show in Table~\ref{app-test_intent}). If previous failures occurred, the Planner incorporates reflection feedback $r$ to avoid repeating the same errors (Lines 4--5).

\noindent\textbf{Execution.} The Executor generates actions $a_t$ based on the current subtask $s$, observation $o_t$, and trajectory $\tau$ (Lines 28--31). Each action is sent to the environment for execution.

\noindent\textbf{Monitoring.} After each action, the Monitor evaluates the execution status $c_t \in \{\texttt{DONE}, \texttt{FAIL}, \texttt{CONTINUE}\}$ by analyzing the environment feedback (Lines 11--20). If the subtask completes successfully (\texttt{DONE}), the system proceeds to the next subtask (Lines 6--10). If a failure is detected (\texttt{FAIL}), control transfers to the Reflector.

\noindent\textbf{Reflection.} The Reflector attributes anomalies to either agent execution errors or GUI defects (Lines 21--27). For agent errors within the retry limit, the Executor attempts self-correction. Otherwise, the system resets and generates a new plan to explore alternative paths.

The loop terminates when all subtasks are completed or the maximum retry limit is reached.

\section{Enabling GUI Agents for Exploratory Testing}\label{appsec:GUIAgents_setup}
Baseline GUI agents are originally designed for navigation tasks without defect detection capabilities. To enable fair evaluation on GUITestBench, we wrap the navigation instructions with explicit testing intent, as shown in Table~\ref{prompt:gui_agent_wrap}.

The wrapped prompt augments the agent's role from a pure navigator to a navigator with testing awareness. It instructs the agent to: (1) adopt a "test engineer" perspective during navigation, monitoring whether each operation produces expected results; (2) report detected defects in a standardized format without interrupting the navigation task. 

To help agents recognize common defect patterns, we provide a checklist covering five categories: incorrect navigation destinations, unresponsive operations, system errors, missing UI elements, and unrelated action results. This checklist is derived from the defect modes in GUITestBench to ensure consistency between the agent's detection scope and the benchmark's evaluation criteria.

\section{Executor Action Space}\label{appsec:action-space}
UI-TARS (7B/72B/1.5-7B)~\cite{UI-TARS} supports \texttt{click}, \texttt{long\_press}, \texttt{type}, \texttt{scroll}, \texttt{open\_app}, \texttt{drag}, \texttt{press\_home}, \texttt{press\_back}, and \texttt{finished}; GUI-Owl (7B/32B)~\cite{mobile-agent-v3} supports \texttt{click}, \texttt{long\_press}, \texttt{swipe}, \texttt{type}, \texttt{answer}, \texttt{system\_button}, \texttt{wait}, and \texttt{terminate}; MAI-UI (8B)~\cite{MAIUI} supports \texttt{click}, \texttt{long\_press}, \texttt{swipe}, \texttt{type}, \texttt{open}, \texttt{drag}, \texttt{answer}, \texttt{system\_button}, \texttt{wait}, and \texttt{terminate}.

\section{Detection Failure of GUI Agents}\label{appsec:other_failure}
\begin{figure*}[tp]
    \centering
    \includegraphics[width=\textwidth]{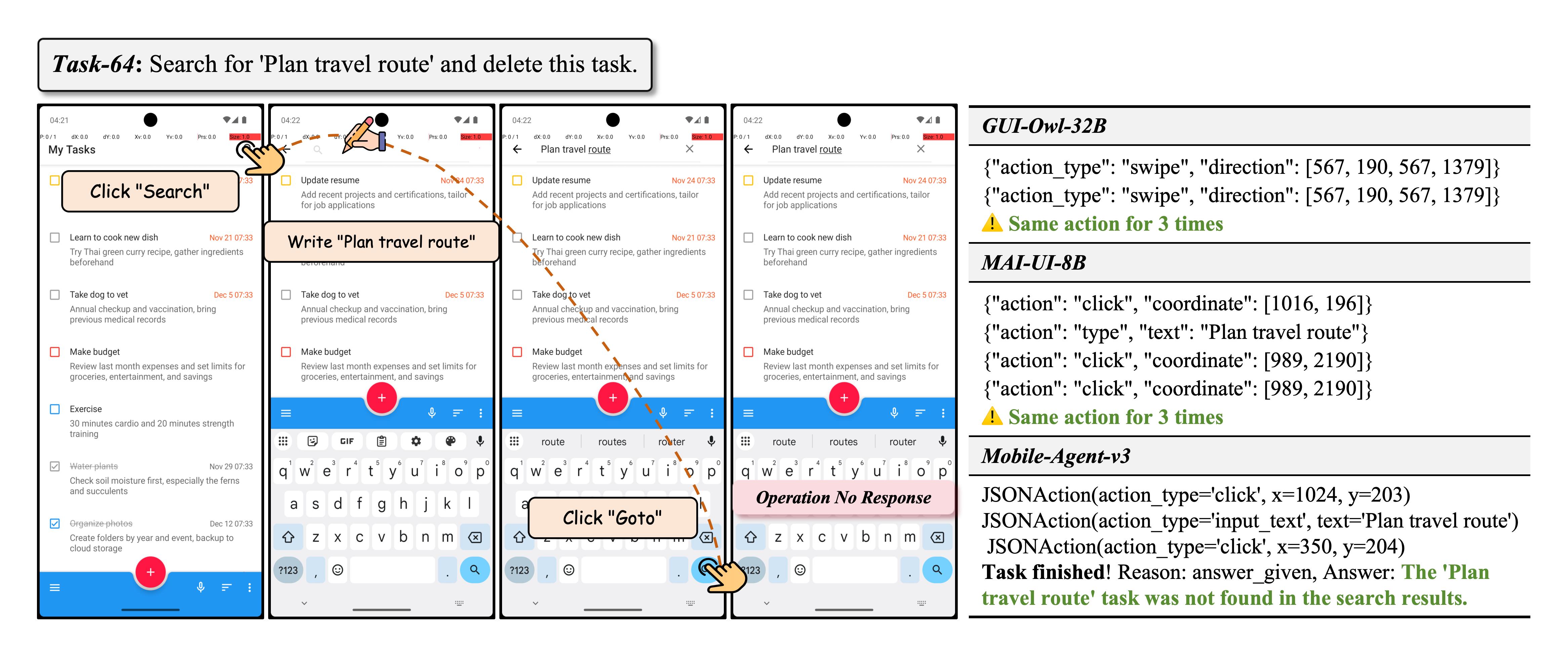}
    \caption{Failure case on Task-64: GUI agents fail to detect an ONR defect in the Tasks app. The task requires searching for "Plan travel route" and deleting it. After navigating to the search results, clicking the "Goto" button produces no response. GUI-Owl-32B repeatedly executes the same swipe action three times without recognizing the anomaly. MAI-UI-8B clicks the target area multiple times but similarly fails to identify the non-responsive state. The preconditions of Mobile-Agent-V3 enabled the model to reach the defect location, but relying solely on the correctness of the trajectory while ignoring environmental feedback caused the model to prematurely declare the task a success.}
    \label{fig:other_failure_cases_01}
\end{figure*}

\begin{figure*}[tp]
    \centering
    \includegraphics[width=\textwidth]{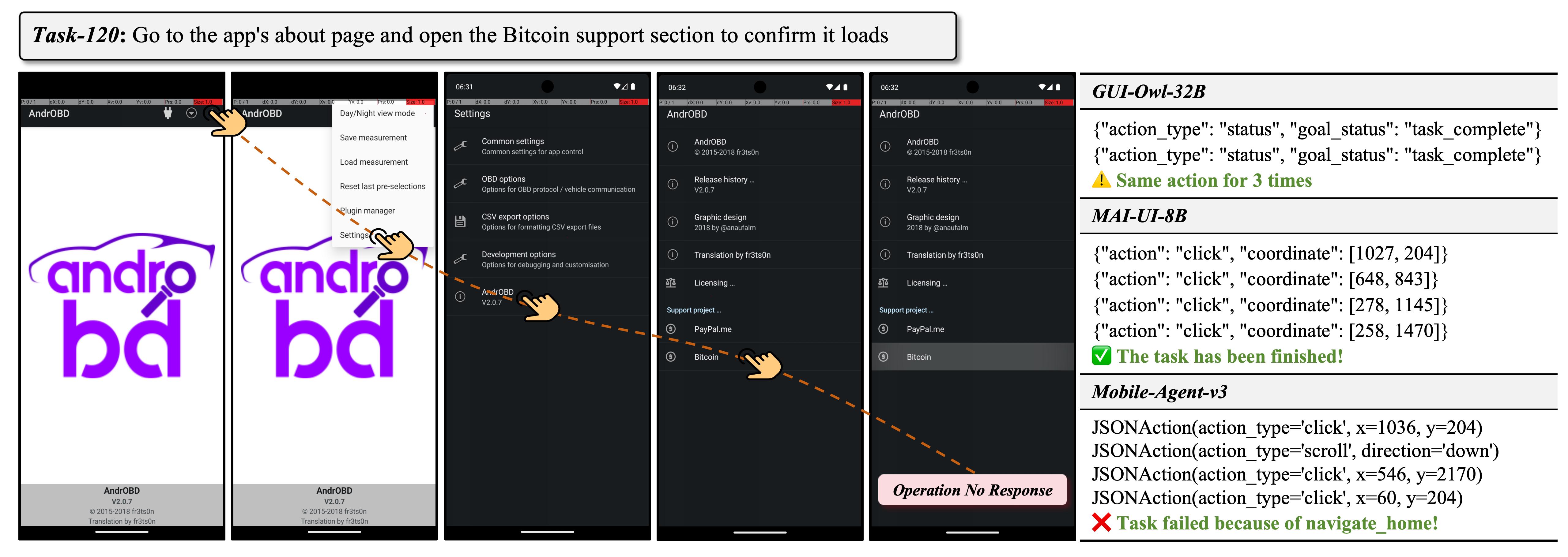}
    \caption{Failure case on Task-120: GUI agents fail to detect an ONR defect in the AndrOBD app. The task requires opening the Bitcoin support section to confirm it loads. When clicking the "Bitcoin" option, the interface fails to respond. GUI-Owl-32B immediately reports "task\_complete" after only two attempts, without verifying whether the page actually loaded. MAI-UI-8B successfully navigates to the target location and happens to trigger the defect, but reports task completion rather than identifying the non-responsive state as an anomaly. Mobile-Agent-V3 attempts multiple clicks but ultimately fails due to an unrelated "navigate\_home" action, completely missing the defect.}
    \label{fig:other_failure_cases_02}
\end{figure*}

We analyze representative failure cases to understand why existing GUI agents fail to detect defects even when navigating close to defect locations (Figure~\ref{fig:other_failure_cases_01} and \ref{fig:other_failure_cases_02}).

\noindent\textbf{Repetition-Induced Termination.} GUI-Owl~\cite{mobile-agent-v3} and MAI-UI~\cite{MAIUI} lack explicit anomaly detection mechanisms. When encountering non-responsive elements, they repeatedly attempt identical actions until the system's termination rule triggers task failure. Without active state verification, these agents cannot distinguish between "action not yet effective" and "action will never be effective due to a defect".

\noindent\textbf{Goal Conflict in Navigation-Oriented Workflows.} Mobile-Agent-V3~\cite{mobile-agent-v3} employs planning, execution, and reflection modules all optimized for navigation success. When repurposed for testing, this creates fundamental conflicts: in Task-64, the agent misinterprets the defect as "task not found"; in Task-120, it triggers an unrelated \texttt{navigate\_home} action. This misalignment causes either premature success or failure declarations without defect reporting.

\section{Detection Failure of GUITester}\label{appsec:failure}
\begin{figure*}[tp]
    \centering
    \includegraphics[width=1.0\linewidth]{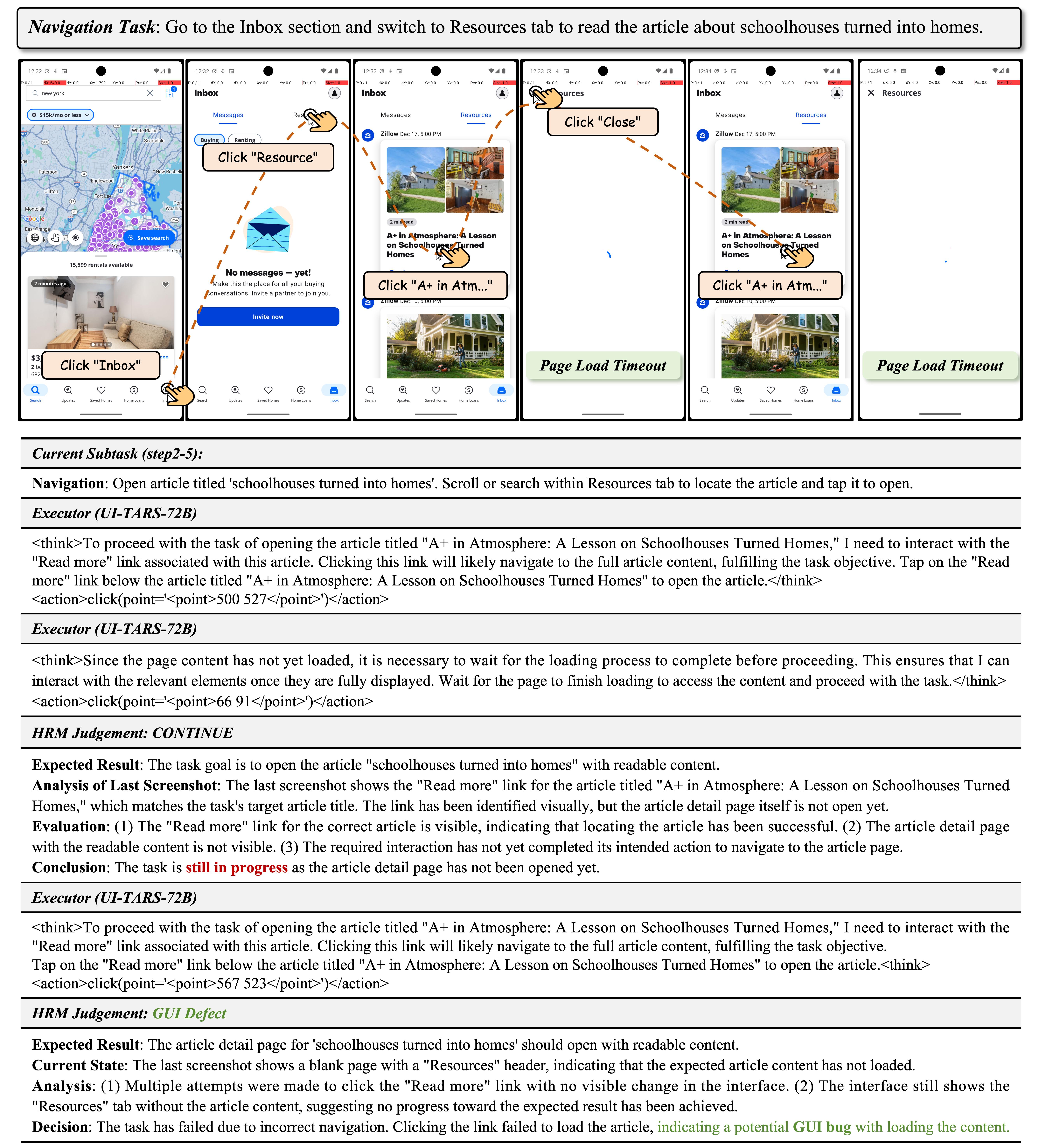}
    \caption{A false positive case caused by premature timeout judgment. GUITester attempts to open an article in the Zillow app, but the page fails to load due to network latency. After the first timeout, HRM issues a CONTINUE state and the Executor retries. When the second attempt also times out, HRM reports it as a GUI defect, despite the actual cause being environmental delay rather than a genuine software defect.}
    \label{fig:failure_cases_01}
\end{figure*}

While GUITester significantly improves defect detection, we identify two primary failure patterns that suggest directions for future improvement:

\noindent\textbf{Premature Timeout Judgment.} The Monitor may misjudge slow-loading states as defects (ONR or UTR) when the environment response time exceeds expectations. As shown in Figure~\ref{fig:failure_cases_01}, GUITester attempts to open an article in the Zillow app. The page is actually loadable, but fails to fully render within the preset 3-second response buffer. As a result, the screenshot captures the interface in a loading state rather than the final content. After the first timeout, HRM correctly issues a CONTINUE state, allowing the Executor to retry. However, when the second attempt also fails to load within the buffer time, HRM concludes that the link is non-responsive and reports it as a GUI defect. This pattern suggests that incorporating adaptive waiting mechanisms or environment-aware timeout thresholds could reduce such false positives.

\begin{figure*}[tp]
    \centering
    \includegraphics[width=1.0\linewidth]{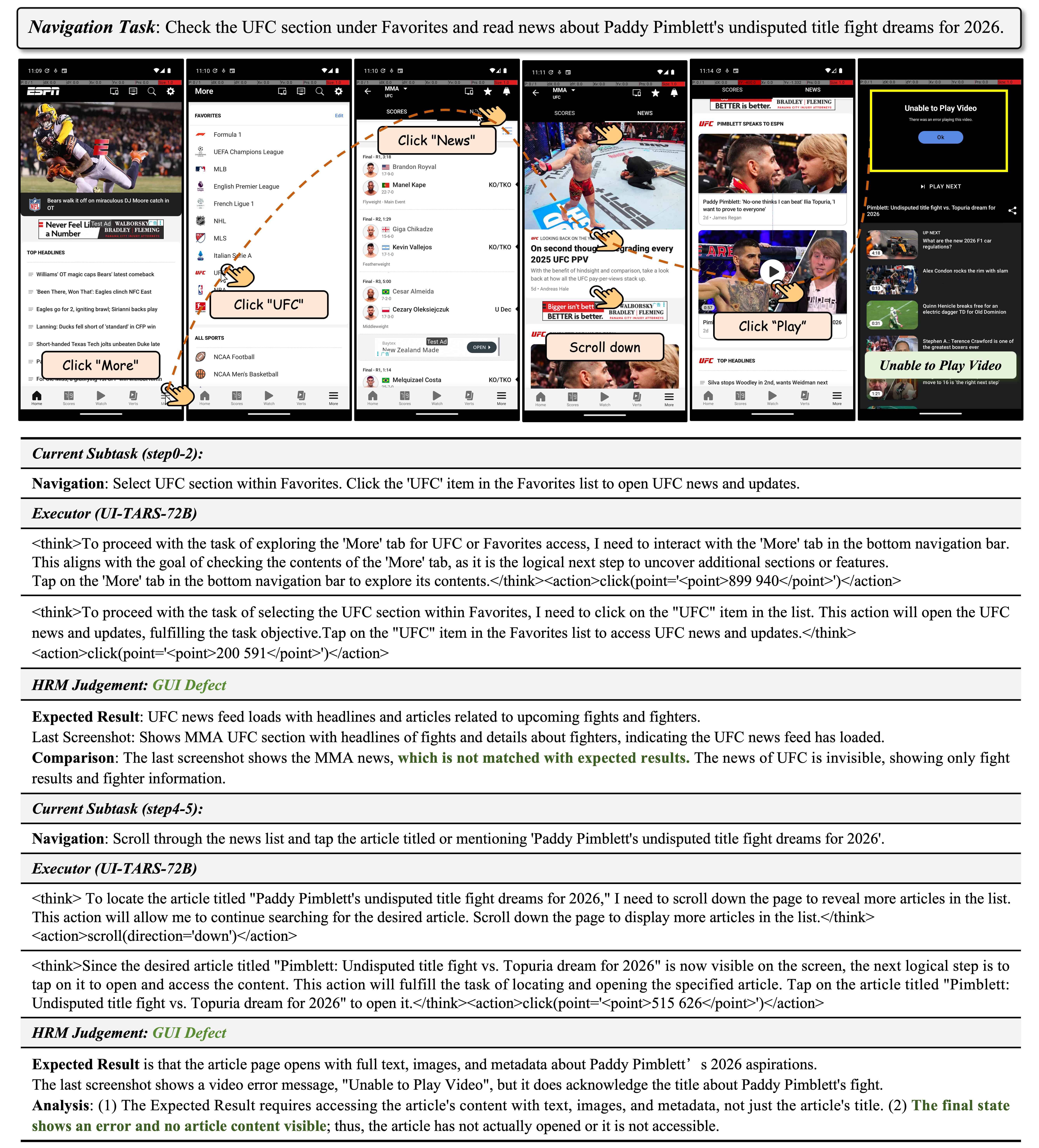}
    \caption{False positive cases caused by Monitor prediction errors. GUITester reports two defects in the ESPN app during a single navigation task: (1) clicking "UFC" navigates to "MMA" section, which HRM misjudges as a navigation logic error due to unfamiliarity with the UFC-MMA relationship; (2) the video page lacks article content, which HRM incorrectly expects based on its assumption about typical video page layouts. Both cases stem from the Monitor's limited domain knowledge rather than genuine GUI defects.}
    \label{fig:failure_cases_02}
\end{figure*}

\noindent\textbf{Monitor Prediction Errors.} Accurate defect detection requires the Monitor to predict expected states after each action. When the Monitor's prediction diverges from the actual expected behavior due to its limited understanding of the application rather than a genuine GUI defect, false positives may occur. As shown in Figure~\ref{fig:failure_cases_02}, GUITester reports two false defects during a single navigation task in the ESPN app. In the first case, clicking "UFC" navigates to the "MMA" section. Since UFC is a subsidiary of MMA, this navigation is correct, but HRM lacks the domain knowledge to recognize this relationship and misjudges it as a navigation logic error. In the second case, HRM expects the video page to contain accompanying article content, text descriptions, or metadata based on its assumptions about typical video page layouts. When the actual page displays only the video player with an error message, HRM reports this mismatch as a defect. Both cases illustrate that the Monitor's effectiveness is bounded by its domain knowledge and assumptions about application behavior. This limitation suggests that integrating application-specific knowledge or more capable vision-language models could improve precision.

\section{More Cases on Released APPs}\label{appsec:more-cases}
\subsection{UI, Unexpected Task result}
\begin{figure*}[tp]
    \centering
    \includegraphics[width=1.0\linewidth]{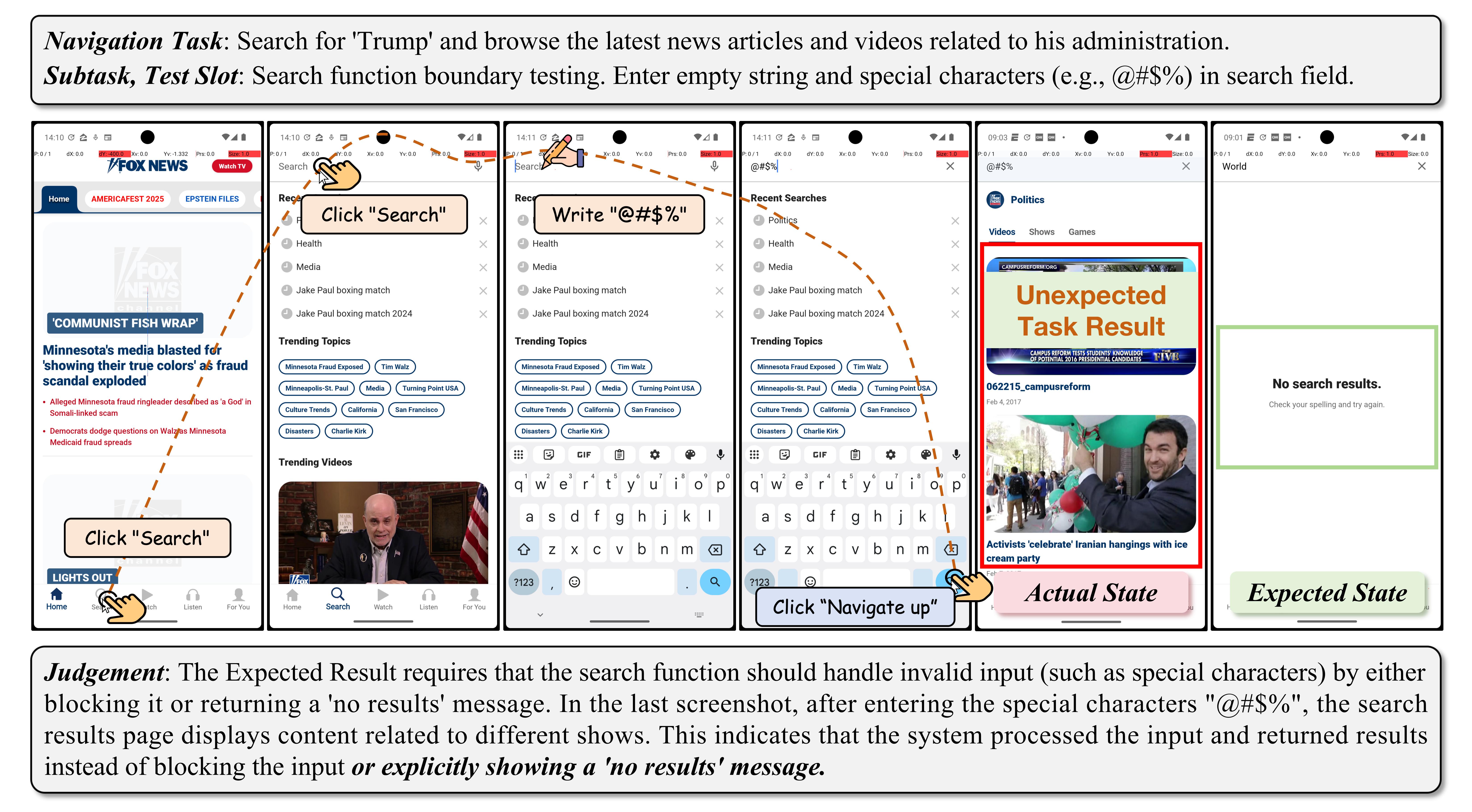}
    \caption{A defect detected by GUITester on Fox News (v5.17.2, December 2025). When entering special characters "@\#\$\%" in the search field, the app returns unrelated content instead of displaying a "no results" message or blocking the invalid input.}
    \label{fig:guitester-fox_news}
\end{figure*}

Figure~\ref{fig:guitester-fox_news} shows a UTR defect discovered on Fox News app. PEM generates a test intent to probe the search function's boundary behavior by entering special characters "@\#\$\%". The search function should either block such input or return a "no results" message; instead, the app accepts it as valid and returns unrelated content about "Politics". This defect would likely be missed by navigation-oriented exploration, which typically uses meaningful search queries rather than edge-case inputs.

\subsection{UX, Unexpected Task result}
\begin{figure*}[tp]
    \centering
    \includegraphics[width=1.0\linewidth]{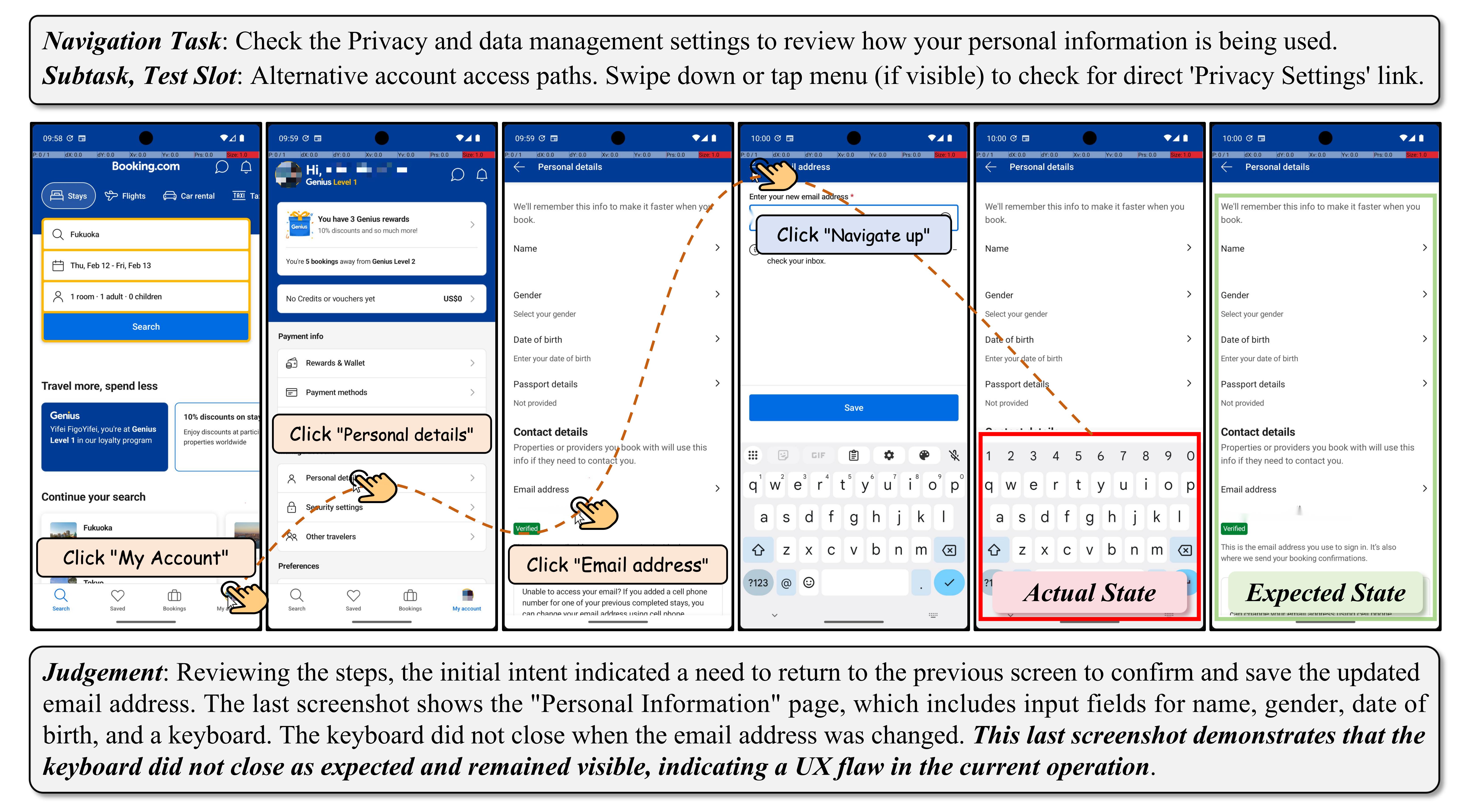}
    \caption{A defect detected by GUITester on Booking.com (v62.1.2, December 2025). After editing the email address in Personal details, the keyboard remains visible when navigating back to the previous screen, instead of automatically dismissing as expected.}
    \label{fig:guitester-booking}
\end{figure*}

Figure~\ref{fig:guitester-booking} shows a UX defect discovered on Booking.com app. After editing the email address and navigating back, the keyboard remains visible instead of automatically dismissing. Unlike functional defects with clear error signals, this UX flaw involves improper state transition that does not block task completion but degrades user experience. The Monitor captures this by detecting the mismatch between expected and actual interface states, demonstrating HRM's sensitivity to subtle interaction anomalies.

\subsection{UI, Navigation Logic Error}
\begin{figure*}[tp]
    \centering
    \includegraphics[width=1.0\linewidth]{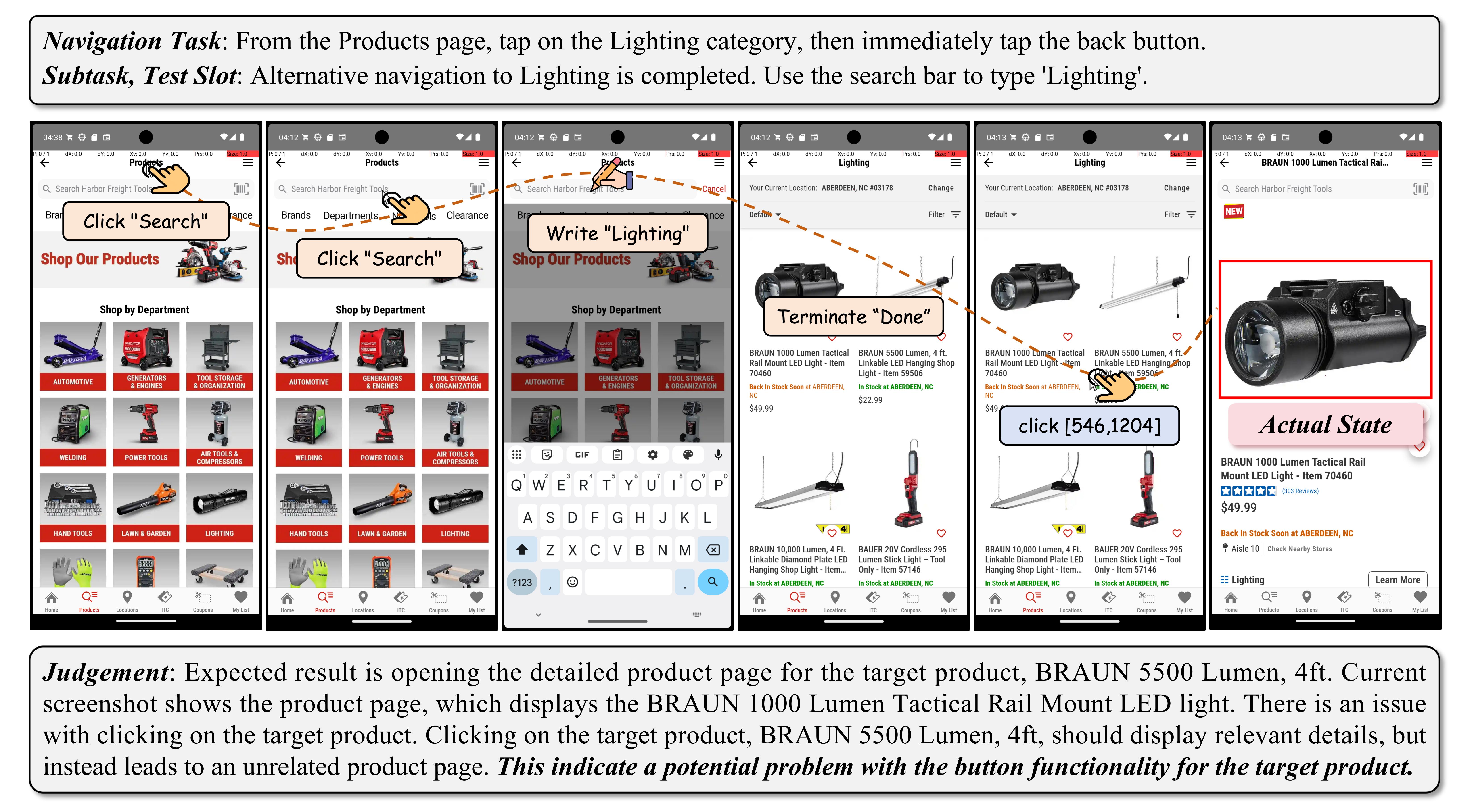}
    \caption{A defect detected by GUITester on Harbor Freight (v2.5.1, December 2025). Clicking on "BRAUN 5500 Lumen, 4ft" product navigates to an unrelated product page displaying "BRAUN 1000 Lumen Tactical Rail Mount LED Light" instead.}
    \label{fig:guitester-harborfreight}
\end{figure*}

Figure~\ref{fig:guitester-harborfreight} shows a NLE defect discovered on Harbor Freight app. When clicking on "BRAUN 5500 Lumen, 4ft" in the product list, the app navigates to an unrelated product page displaying "BRAUN 1000 Lumen Tactical Rail Mount LED Light". The Monitor detects this navigation logic error by identifying the mismatch between the clicked element and the resulting page content.

\subsection{UX, Navigation Logic Error}
\begin{figure*}[tp]
    \centering
    \includegraphics[width=1.0\linewidth]{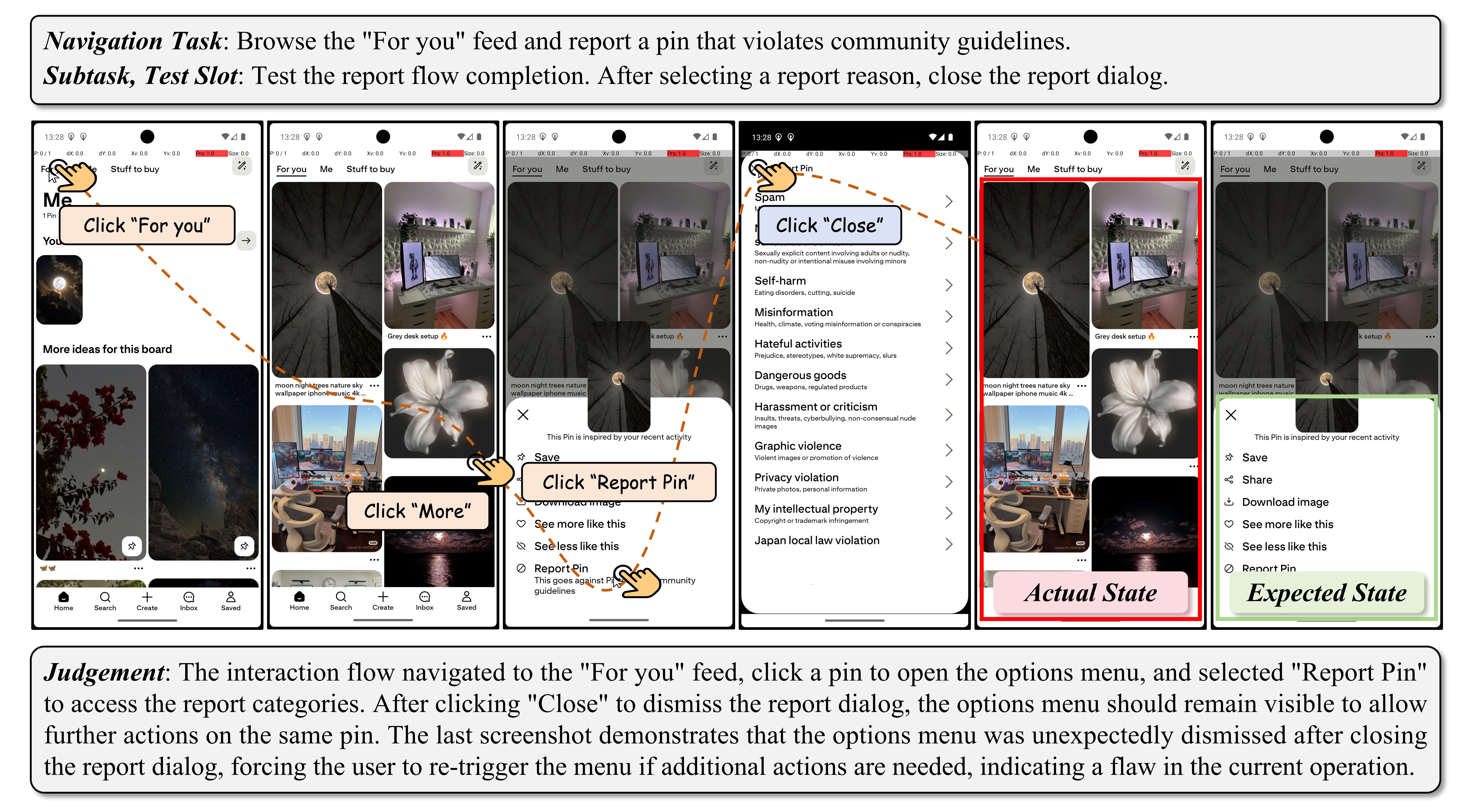}
    \caption{A defect detected by GUITester on Pinterest (v13.46.1, December 2025). After opening the "Report Pin" dialog and clicking "Close" to dismiss it, the options menu is unexpectedly closed instead of remaining visible for further actions on the same pin.}
    \label{fig:guitester-pinterest}
\end{figure*}

Figure~\ref{fig:guitester-pinterest} shows a UX defect discovered on Pinterest app. After opening the "Report Pin" dialog and clicking "Close" to dismiss it, the options menu is unexpectedly closed instead of remaining visible for further actions. Unlike functional defects with explicit error messages, this UX flaw involves unexpected state restoration that does not prevent task completion but disrupts the natural interaction flow. The Monitor captures this by detecting the mismatch between the expected state (options menu visible) and the actual state (menu dismissed).

\begin{table*}[!ht]
    \small
    \centering
    \begin{tcolorbox}[colframe=black, colback=gray!10!white, coltitle=black, boxrule=0.5mm]
        You are a GUI Defect Verification Expert. Your task is to determine whether the agent's exploration trajectory has successfully reproduced a known GUI defect. \\
        You'll receive the following information:
        \begin{itemize}[leftmargin=1em,itemsep=-1pt]
            \item \textbf{Defect Description}: A detailed description of the known GUI defect, including its preconditions, trigger action, and expected result.
            \item \textbf{Defect Behavior Reference Images}: Screenshots showing the defect behavior.
            \item \textbf{Agent's Trajectory}: A sequence of actions and final observations performed by the agent.
            \item \textbf{Agent's Trajectory Images}: Screenshots showing the agent's final state.
        \end{itemize}
        \# \textbf{Your Task}\\
        Analyze the agent's trajectory and determine whether the defect was successfully reproduced.\\
        \#\# \textbf{Verification Checklist}\\
        You MUST explicitly verify each of the following:
        \begin{enumerate}[leftmargin=1em,itemsep=-1pt]
        \item \textbf{Precondition Check}: Did the agent correctly establish ALL preconditions?
        \item \textbf{Trigger Action Check}: Did the agent perform the correct trigger action?
        \item \textbf{Result Check}: Does the final state match the known defect behavior?
        \item \textbf{Final Verdict}: 
            \begin{enumerate}[leftmargin=1em,itemsep=-1pt]
                \item \textbf{GUI\_BUG}: Agent successfully reproduced the defect.
                \item \textbf{EXECUTOR\_ERROR}: The agent failed to reproduce the defect.
            \end{enumerate}
        \end{enumerate}
        \#\# \textbf{Important Note}
        \begin{enumerate}[leftmargin=1em,itemsep=-1pt]
            \item \textbf{Action Sequence Matters}: A trajectory that only sets up preconditions but MISSES the trigger action should be marked as \textbf{EXECUTOR\_ERROR}. The trigger action is the critical step that actually causes the defect to manifest.
            \item \textbf{Input Values May Vary}: The defect description uses specific examples (e.g., "Pay Bills") for illustration purposes. The agent may use different input values (e.g., "XYZ123"), which is acceptable. However, the \textbf{action sequence and interaction pattern} must match the defect description exactly.
            \item \textbf{Be Strict About Completeness}: Do not assume or infer actions that are not explicitly documented in the trajectory. If a required action is missing from the trajectory, it was not performed.
        \end{enumerate}
        \# \textbf{Output Format}
        \begin{enumerate}[leftmargin=1em,itemsep=-1pt]
            \item Please provide a clear and concise thought within \texttt{<think>} and \texttt{</think>} tags.
            \item Please provide the final verdict within \texttt{<answer>} and \texttt{</answer>} tags.
            \item The final verdict must be either \textbf{GUI\_BUG} or \textbf{EXECUTOR\_ERROR}.
            \item Here is a sample output: \texttt{<think>}your thought\texttt{</think><answer>}final verdict\texttt{</answer>}.
        \end{enumerate}
    \end{tcolorbox}
    \caption{Judge Model Prompt}
    \label{prompt:judge_model}
\end{table*}

\begin{table*}[!t]
    \small
    \centering
    \begin{tcolorbox}[colframe=black, colback=gray!10!white, coltitle=black, boxrule=0.5mm]
        You are a GUI Agent with defect detection capabilities. While completing the navigation task, you need to explore and examine each interface from the perspective of a "test engineer":
        \begin{itemize}[leftmargin=1em,itemsep=-1pt]
            \item [-] Before and after each execution: Consider whether the model's response matches expectations.
            \item [-] Throughout: Pay attention to anything that seems "off".
        \end{itemize}
        \textbf{Bug Reporting Requirements}
        \begin{itemize}[leftmargin=1em,itemsep=-1pt]
            \item [-] When you find a potential defect, output it as \{"action": "answer", "text": "GUI\_BUG"\}.
            \item [-] Do not interrupt the main task because of finding a defect.
        \end{itemize}
        The name of current app is \{\texttt{app\_name}\}.\\
        The user query: \{\texttt{instruction}\}.\\
        Task progress (You have done the following operation on the current device): \{\texttt{histories}\}.\\
        
        \textbf{The following tips can help you complete user tasks}: 
        \begin{enumerate}[leftmargin=1em,itemsep=-1pt]
            \item Wrong Destination / Incorrect Navigation
            \begin{itemize}[leftmargin=1em,itemsep=-1pt]
                \item [-] Action intended to open/navigate to X, but system opened/navigated to Y instead.
            \end{itemize}
            \item Action Had No Effect After Multiple Attempts
            \begin{itemize}[leftmargin=1em,itemsep=-1pt]
                \item [-] The same action was repeated 2+ times with no state change.
                \item [-] The UI remains completely unchanged despite valid interaction attempts.
            \end{itemize}
            \item Error State or System Failure
            \begin{itemize}[leftmargin=1em,itemsep=-1pt]
                \item [-] An error message appeared and blocked progress.
                \item [-] The interface crashed, froze, or entered an unrecoverable state.
                \item [-] The app closed unexpectedly.
            \end{itemize}
            \item Required Element Permanently Missing
            \begin{itemize}[leftmargin=1em,itemsep=-1pt]
                \item [-] The target UI element does not exist and cannot be accessed through any reasonable path.
                \item [-] The feature appears to be unimplemented or broken.
            \end{itemize}
            \item Completely Unrelated Result
            \begin{itemize}
                \item [-] The action triggered functionality that is logically unrelated to the intent.
            \end{itemize}
        \end{enumerate}
    \end{tcolorbox}
    \caption{Exploratory GUI Testing Wrapped Prompt}
    \label{prompt:gui_agent_wrap}
\end{table*}

\end{document}

%% file: table-overall.tex
\begin{table*}[t]
    \centering
    \footnotesize
    \setlength{\tabcolsep}{3pt}
    \renewcommand{\arraystretch}{1.05}
    \begin{tabular}{l|cccccc|cccc|cc}
        \toprule
        \multirow{2}{*}{\textbf{Model}} & \multicolumn{2}{c}{\textbf{UI-ONR}} & \multicolumn{2}{c}{\textbf{UI-UTR}} & \multicolumn{2}{c}{\textbf{UI-NLE}} & \multicolumn{2}{c}{\textbf{UX-UTR}} & \multicolumn{2}{c}{\textbf{UX-NLE}} & \multicolumn{2}{c}{\textbf{Overall}} \\
        & Recall$\uparrow$ & F1$\uparrow$ & Recall$\uparrow$ & F1$\uparrow$ & Recall$\uparrow$ & F1$\uparrow$ & Recall$\uparrow$ & F1$\uparrow$ & Recall$\uparrow$ & F1$\uparrow$ & Recall$\uparrow$ & F1$\uparrow$ \\
        \midrule
        \rowcolor{gray!20}
        \textit{\textbf{Pass@1}} & & & & & & & & & & & & \\
        GUI-Owl-7B & 0.00 & 0.00 & 0.00 & 0.00 & 0.00 & 0.00 & 0.00 & 0.00 & 0.00 & 0.00 & 0.00 & 0.00 \\
        GUI-Owl-32B & 1.70 & 3.10 & 3.10 & 5.60 & 6.50 & 11.80 & 0.00 & 0.00 & 0.00 & 0.00 & 2.80 & 5.10 \\
        MAI-UI-8B & 1.13 & 2.13 & 6.23 & 11.60 & 0.00 & 0.00 & 0.00 & 0.00 & 0.00 & 0.00 & 1.87 & 3.60 \\
        UI-TARS-7B & 1.70 & 3.30 & 0.00 & 0.00 & 3.20 & 5.30 & 0.00 & 0.00 & 0.00 & 0.00 & 1.40 & 2.60 \\
        UI-TARS-72B & 16.10 & 23.65 & 9.45 & 15.00 & 3.23 & 5.27 & 0.00 & 0.00 & 2.08 & 2.78 & 9.68 & 15.03 \\
        UI-TARS-1.5-7B & 22.25 & 28.13 & 16.68 & 23.20 & 9.70 & 15.33 & 9.38 & \underline{12.95} & \underline{12.53} & \underline{18.38} & 16.73 & 22.95 \\
        \midrule
        Mobile-Agent-V3 & 0.00 & 0.00 & 1.03 & 2.03 & 3.20 & 6.00 & 0.00 & 0.00 & 0.00 & 0.00 & 0.93 & 1.83 \\
        \midrule
        $\text{GUITester}_{\;\textbf{\textit{(GUI-Owl-32B)}}}$ & 16.25 & 18.45 & 15.65 & 17.30 & 6.45 & 7.40 & 0.00 & 0.00 & 0.00 & 0.00 & 11.80 & 13.55 \\
        $\text{GUITester}_{\;\textbf{\textit{(UI-TARS-72B)}}}$ & 21.34 & 23.94 & 31.86 & 35.88 & \underline{47.74} & \underline{49.96} & 0.00 & 0.00 & 8.30 & 9.50 & 26.60 & 29.50 \\
        $\text{GUITester}_{\;\textbf{\textit{(UI-TARS-1.5-7B)}}}$ & \underline{26.68} & \underline{30.18} & \underline{34.38} & \underline{39.10} & 39.35 & 44.93 & \underline{12.50} & 12.50 & 12.50 & 16.70 & \underline{28.10} & \underline{32.18} \\
        \midrule
        \rowcolor{gray!20}
        \textit{\textbf{Pass@3}} & & & & & & & & & & & & \\
        UI-TARS-72B & 28.30 & 38.20 & 16.10 & 21.95 & 9.70 & 13.80 & 6.25 & 6.65 & 0.00 & 0.00 & 17.95 & 19.55 \\
        UI-TARS-1.5-7B & 35.60 & 42.00 & 29.00 & 34.95 & 16.10 & 22.45 & 12.50 & \textbf{14.30} & 16.70 & 22.20 & 26.95 & 33.35 \\
        \midrule
        
        $\text{GUITester}_{\;\textbf{\textit{(GUI-Owl-32B)}}}$ & 28.60 & 29.30 & 28.10 & 29.00 & 12.90 & 13.30 & 0.00 & 0.00 & 0.00 & 0.00 & 21.40 & 22.10 \\
        
        $\text{GUITester}_{\;\textbf{\textit{(UI-TARS-72B)}}}$ & 40.00 & 40.35 & 43.80 & 43.80 & \textbf{75.80} & \textbf{75.80} & 0.00 & 0.00 & 8.30 & 9.50 & 43.40 & 43.70 \\
        $\text{GUITester}_{\;\textbf{\textit{(UI-TARS-1.5-7B)}}}$ & \textbf{45.00} & \textbf{46.60} & \textbf{50.00} & \textbf{50.00} & 70.00 & 71.20 & \textbf{12.50} & 12.50 & \textbf{25.00} & \textbf{26.10} & \textbf{47.90} & \textbf{48.90} \\
        \bottomrule
    \end{tabular}
    \caption{GUI defect detection results on GUITestBench. \textbf{Bold} and \underline{underlined} numbers indicate the best scores under the Pass@3 and Pass@1 settings, respectively. Since the improvements of the GUI-Owl, MAI-UI-8B and UI-TARS-7B on Pass@3 are not significant, we have not reported the corresponding results.}
    \label{tab:main_results}
\end{table*}

%% file: table-analysis.tex
\begin{table*}[!t]
    \centering
    \footnotesize
    \setlength{\tabcolsep}{5pt}
    \renewcommand{\arraystretch}{1.05}
    \begin{tabular}{l|cccc|cccc}
        \toprule
        \multirow{2}{*}{\textbf{Model}} & \multicolumn{2}{c}{\textbf{Defect-Ori}} & \multicolumn{2}{c|}{\textbf{Explore-Ori}} & \multicolumn{2}{c}{\textbf{Single-Act}} & \multicolumn{2}{c}{\textbf{Multi-Act}} \\
        & Recall$\uparrow$ & F1$\uparrow$ & Recall$\uparrow$ & F1$\uparrow$ & Recall$\uparrow$ & F1$\uparrow$ & Recall$\uparrow$ & F1$\uparrow$ \\
        \midrule
        GUI-Owl-7B & 0.00 & 0.00 & 0.00 & 0.00 & 0.00 & 0.00 & 0.00 & 0.00 \\
        GUI-Owl-32B & 5.60 & 10.30 & 1.90 & 3.40 & 4.30 & 7.70 & 0.00 & 0.00 \\
        MAI-UI-8B & 3.60 & 6.37 & 1.53 & 3.00 & 0.00 & 0.00 & 3.77 & 7.00 \\
        UI-TARS-7B & 2.80 & 5.10 & 1.00 & 1.80 & 1.10 & 2.10 & 1.90 & 3.60 \\
        UI-TARS-72B & 16.70 & 26.70 & 10.30 & 16.40 & 13.50 & 21.10 & 9.30 & 15.40 \\
        UI-TARS-1.5-7B & 22.90 & 23.90 & 14.30 & 20.40 & 20.20 & 28.30 & 15.40 & 20.50 \\
        \midrule
        Mobile-Agent-V3 & 0.00 & 0.00 & 1.23 & 2.40 & 0.00 & 0.00 & 1.43 & 2.77 \\
        \midrule
        $\text{GUITester}_{\;\textbf{\textit{(GUI-Owl-32B)}}}$ & 19.40 & 21.50 & 9.40 & 10.90 & 16.50 & 18.90 & 6.20 & 7.00 \\
        $\text{GUITester}_{\;\textbf{\textit{(UI-TARS-72B)}}}$ & \textbf{33.30} & \textbf{35.80} & \underline{29.90} & \underline{33.30} & \textbf{48.30} & \textbf{51.50} & \underline{17.90} & \underline{21.20} \\
        $\text{GUITester}_{\;\textbf{\textit{(UI-TARS-1.5-7B)}}}$ & \underline{30.60} & \underline{34.40} & \textbf{28.00} & \textbf{31.20} & \underline{43.80} & \underline{47.00} & \textbf{18.75} & \textbf{23.40} \\
        \bottomrule
    \end{tabular}
    \caption{Results across task types and defect complexity.}
    \label{tab:main_results-2}
\end{table*}